\documentclass[sigconf,nonacm]{aamas} %

\usepackage{balance} %

\setcopyright{none} %

\newcommand{\method}{DUOS}
\title{Discovering Update Rules for Opponent Shaping\texorpdfstring{\\[-2.2pt]}{ }in General-Sum Games}

\author{Hong Jun Lee}
\affiliation{
  \institution{King's College London}
  \city{London}
  \country{United Kingdom}}
\email{hong\_jun.lee@kcl.ac.uk}

\author{Ziyan Wang}
\affiliation{
  \institution{Harvard University}
  \city{Cambridge}
  \state{MA}
  \country{United States}}
\email{ziyan\_wang@seas.harvard.edu}

\author{Jun Wang}
\affiliation{
  \institution{University College London}
  \city{London}
  \country{United Kingdom}}
\email{junwang@cs.ucl.ac.uk}

\author{Stefanos Leonardos}
\affiliation{
  \institution{King's College London}
  \city{London}
  \country{United Kingdom}}
\email{stefanos.leonardos@kcl.ac.uk}

\author{Yali Du}
\affiliation{
  \institution{King's College London}
  \city{London}
  \country{United Kingdom}}
\email{yali.du@kcl.ac.uk}

\begin{abstract}
In general-sum games, self-interested learners can reach outcomes that are poor for everyone, which has motivated methods that shape the opponent's learning. Methods like Learning with Opponent-Learning Awareness (LOLA), which differentiates through the opponent's learning process, and Model-Free Opponent Shaping (M-FOS), which uses model-free optimisation methods for long-horizon opponent shaping, have enabled mutual cooperation and exploitation of learners. However, these methods require access to the opponent's parameters, policy, rewards or learning rule during deployment. In single-agent reinforcement learning, update rules discovered by meta-learning have achieved state-of-the-art results (e.g. DiscoRL). These rules are separate from the policy they train and read only the agent's own experience. We hereby extend this framework to general-sum settings by introducing Discovered Update Rule for Opponent Shaping (\method{}). To the best of our knowledge, this is the first method to discover an update rule for an agent in general-sum games. \method{} meta-trains the rule to maximise the agent's expected return over its lifetime, and updates an inner network using the agent's own experience. We show that our discovered rule exhibits behaviour consistent with opponent shaping without reading opponent policies or parameters, and ranks \textit{1st} in normalised mean score and \textit{1st} in self-play among tournament entrants.
\end{abstract}


\newcommand{\BibTeX}{\rm B\kern-.05em{\sc i\kern-.025em b}\kern-.08em\TeX}
\usepackage{placeins} %
\AtBeginDocument{\setcounter{dbltopnumber}{2}} %
\AtBeginDocument{\expandafter\def\expandafter\UrlOrds\expandafter{\UrlOrds\do\:}\expandafter\def\expandafter\UrlSpecials\expandafter{\UrlSpecials\do\/{\mathchar"2F\futurelet\URLnext\URLslash}}}\def\URLslash{\ifx\URLnext/\else\penalty\UrlBreakPenalty\fi}%
\ifdefined\footins\AtBeginDocument{\skip\footins=7pt}\fi\usepackage{adjustbox} %
\usepackage{algorithm}
\usepackage[noend]{algpseudocode}
\usepackage{booktabs, float}

\begin{document}

\fancypagestyle{firstpagestyle}{\fancyhf{}\renewcommand{\headrulewidth}{0pt}\renewcommand{\footrulewidth}{0pt}\fancyfoot[C]{\footnotesize\thepage}}
\fancypagestyle{standardpagestyle}{\fancyhf{}\renewcommand{\headrulewidth}{0pt}\renewcommand{\footrulewidth}{0pt}\fancyfoot[C]{\footnotesize\thepage}}
\pagestyle{standardpagestyle}

\maketitle 
\begingroup
\renewcommand{\thefootnote}{}%
\footnotetext{Preprint. Under review.}%
\endgroup

\section{Introduction}

Real-world multi-agent environments, from self-driving cars \cite{schwarting_social_2019} to multi-agent large language model (LLM) systems \cite{park_generative_2023,akata_playing_2025}, can be framed as games ranging from zero-sum to fully cooperative. General-sum games cover the mixed-motive settings in between, such as social dilemmas \cite{dawes_social_1980, leibo_multi-agent_2017} or buyer-seller negotiations \cite{lewis_deal_2017}. Naive self-interested learners within general-sum games often converge towards outcomes that are poor for everyone \cite{foerster_learning_2018,leibo_multi-agent_2017}. For example, in Iterated Prisoner's Dilemma (IPD) \cite{axelrod_evolution_2006,press_iterated_2012}, they often converge to mutual defection rather than mutual cooperation. Opponent shaping \cite{foerster_learning_2018,lu_model-free_2022} addresses this problem by influencing how others learn to achieve mutually beneficial outcomes, although it can also be used to exploit them.

Existing shaping methods leave a gap. Hand-derived rules such as LOLA \cite{foerster_learning_2018} and POLA \cite{zhao_proximal_2022} differentiate through the opponent's update, and require their parameters, payoffs or learning rules. Meta-learned shapers such as M-FOS \cite{lu_model-free_2022}, SHAPER \cite{khan_scaling_2024} and COALA-PG \cite{meulemans_multi-agent_2025} do not require a model of the opponent's learning rule as SHAPER and COALA-PG act from their own history, but their learning algorithm is inside a black-box network, retrained each game. In single-agent reinforcement learning (RL), where the environment is stationary,  rules discovered by meta-learning \cite{oh_discovering_2020,oh_discovering_2025}, rather than designed by hand, have achieved state-of-the-art results. Such rules also have practical advantages over a black-box learner: a discovered rule is not limited to updates derived by hand \cite{oh_discovering_2020}, is separate from the policy it trains, and can transfer to different games without retraining. We therefore ask: can an update rule that learns from an agent's own experience be discovered for general-sum games? However, learning update rules is difficult because learning is coupled, meta-gradients explode through co-learning, and opponents adapt, making the setting non-stationary. We hereby introduce \textbf{D}iscovered \textbf{U}pdate Rule for \textbf{O}pponent \textbf{S}haping (\method{}), a meta-reinforcement learning algorithm that discovers its own update rule by interacting with opponents, without needing their policies or parameters at \mbox{deployment}. \looseness=1{\emergencystretch=4pt\par}

In this paper, we design the meta-objective to maximise the agent's return over its lifetime. The rule reads the agent's own experience and outputs policy updates, hence at deployment it does not require any privileged information: the agent never sees its opponents' parameters, policies, learning rules or rewards. Each lifetime starts from a random policy, so what the agent learns comes from the rule. To avoid meta-gradients that explode through co-learning, the rule is trained with a likelihood-ratio meta-gradient, without differentiating through any learning. We then examine whether opponent shaping emerges from the rule without specifying it explicitly.

Although \method{} is trained to maximise lifetime return, this objective alone does not establish robustness to opponents trained specifically against its learned update rule. We therefore add a stage inspired by Policy Space Response Oracles (PSRO) \citep{lanctot_unified_2017} to meta-training: we train opponents to maximise their returns against the current rule, add them to the training population, and continue optimising the rule against the expanded population. We evaluate whether PSRO improves resistance to exploitation for \method{} and how it affects tournament results.

We evaluate \method{} in IPD, Iterated Matching Pennies (IMP), Chicken, Stag Hunt and Coin Game \citep{lerer_maintaining_2018} in round-robin tournaments against established opponent-shaping methods, including methods that observe their opponents' policies or parameters. Each baseline is reproduced from its original settings, checked against results where possible, and flagged if it fails. We also observe \method{} self-play and zero-shot transfer of the IPD-trained rule to Chicken and Stag Hunt. Despite using only its own experience and learning an update rule from scratch, \method{} ranks \textit{1st} in normalised mean score across all games tested, and \textit{1st} in normalised mean score for self-play among tournament entrants.

\section{Related Work}
\textbf{Opponent shaping:} LOLA introduced the idea of opponent shaping by differentiating through the opponent's learning step \cite{foerster_learning_2018}, which DiCE estimates from sampled trajectories \citep{foerster_dice_2018}. Stable Opponent Shaping \cite{letcher_stable_2019}, COLA \cite{willi_cola_2022} and POLA \cite{zhao_proximal_2022} then further refined the idea, each requiring the opponent's parameters (or in POLA's opponent-modelling variant, a model of its policy fitted to its actions), as well as its payoffs and learning rule. LOQA \cite{aghajohari_loqa_2024} and Advantage Alignment \cite{duque_advantage_2025} instead shape through estimates of the opponent's values, while Reciprocator \cite{zhou_reciprocal_2024} and LIO \cite{yang_learning_2020} shape through rewards tied to the opponent's returns. Best Response Shaping trains against an approximate best response to the agent instead \citep{aghajohari_best_2024}. All of these methods need opponent information each time they update their policy, whereas \method{} only needs it during meta-training.

One level above, meta-game methods learn to shape over many episodes. M-FOS trains a meta-policy where the action is the agent's next policy \cite{lu_model-free_2022}. For matrix games, M-FOS observes both players' policies, and in Coin Game, it only observes the game state and both players' rewards. SHAPER \cite{khan_scaling_2024}, COALA-PG \citep{meulemans_multi-agent_2025}, and in-context co-player inference \cite{weis_multi-agent_2026} use recurrent or sequence-model agents that act with their own history. For all of these methods, the opponent-shaping behaviour and the learning algorithm are held in a black-box network. Meta-MAPG \cite{kim_policy_2021} and FURTHER \cite{kim_influencing_2022} also take into account other agents' learning, through meta-gradients and by influencing their long-run behaviour, respectively. The Good Shepherd \citep{balaguer_good_2022} optimises a fixed policy that shapes learners without learning at deployment. \method{} updates a separate policy, but unlike M-FOS, it reads only its own experience and payoffs.

\textbf{Learned update functions for games:} COLA represents both players' update functions with a neural network, and fits them to a consistency equation derived from LOLA \citep{willi_cola_2022}. Meta-Value Learning (MeVa) learns a value over the joint policy, and the gradient of the value leads to each player's update \citep{cooijmans_meta-value_2023}. MeVa extorts naive learners in IPD but reads both players' parameters and is fitted to each game. 

Update rules for centralised solvers have been discovered by meta-gradients or LLM-driven search \citep{feng_neural_2021,xu_autocfr_2022,li_discovering_2026} for games but not for agents. Other work instead learns optimisers or regret minimisers that converge towards equilibrium \citep{vadori_consensus_2022,harris_meta-learning_2023,sychrovsky_learning_2024,sychrovsky_approximating_2026, chen_learning_2023}. These target consistency per game or equilibrium computation, whereas \method{} is meta-trained for an agent's own lifetime return against a population of opponents, reading only its experience.

\textbf{Discovered reinforcement learning algorithms:} 
In single-agent reinforcement learning, \citet{xu_meta-gradient_2018} use meta-gradients to tune hyperparameters such as the discount factor, and later to discover the objective itself \citep{xu_meta-gradient_2020}. LPG goes further and meta-learns an entire update rule across many lifetimes, generalising from toy environments to Atari \citep{oh_discovering_2020}. DiscoRL \citep{oh_discovering_2025} scales this approach to a large number of environments and discovers a rule that outperforms state-of-the-art hand-designed algorithms. Related work discovers objectives, drift functions and optimisers \citep{kirsch_improving_2020, co-reyes_evolving_2021, lu_discovered_2022, jackson_discovering_2023, jackson_discovering_2024, goldie_can_2024}. Other work analyses the meta-gradients \cite{flennerhag_bootstrapped_2022, vuorio_investigation_2022, flennerhag_optimistic_2023} that these algorithms utilise. Black-box meta-learners such as RL$^2$ \citep{duan_rl_2016} keep the learning algorithm within the agent's network, whereas these methods separate the rule from the agent it trains \citep{kirsch_introducing_2022}. These methods form the many-shot branch of meta-reinforcement learning \citep{beck_tutorial_2025}, where a learning algorithm trains an agent over a long lifetime. \method{} belongs to this branch and is applied to general-sum games, where opponents also learn.

Evolutionary game theory has asked which learning rules survive selection in repeated games, selecting from hand-designed learning rules or their parameters \citep{couto_evolution_2026,dridi_learning_2018,mcavoy_evolutionary_2022}. To our knowledge, \method{} is the first update rule discovered for an agent in general-sum games.

\begin{figure*}[t] \centering \includegraphics[width=\textwidth]{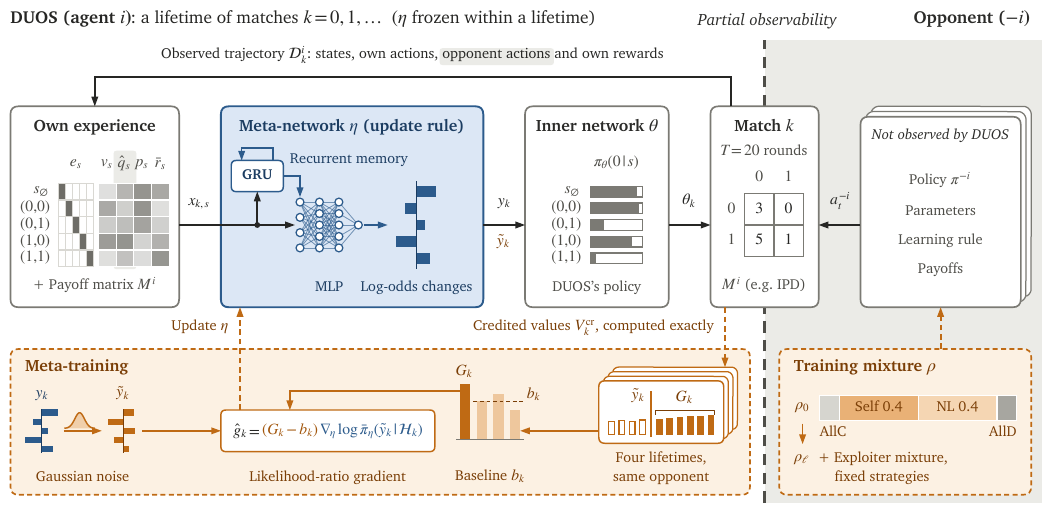} \Description{A diagram of DUOS in three parts. Centre: in each match of a lifetime, the agent's own experience (its states, own and opponent actions, own rewards and its payoff matrix) feeds the meta-network, a GRU and an MLP, which outputs log-odds changes. The inner network is fitted to them and gives the policy for the next match. Right: the opponent, whose policy, parameters, learning rule and payoffs DUOS does not observe. Bottom: meta-training adds Gaussian noise to the log-odds changes and updates the meta-network with a likelihood-ratio gradient whose baseline comes from four lifetimes against the same opponent, drawn from a training mixture of self-play, naive learners, always-cooperate and always-defect, plus exploiters.} \caption{Overview of \method{}. After every match, the meta-network $\eta$ reads the agent's experience $\mathcal D^i_k$ and outputs log-odds changes $y_k$, to which the inner network $\theta$ is fitted. The opponent's policy, parameters, learning rule and payoffs are never observed. Meta-training perturbs $y_k$ with Gaussian noise and updates $\eta$ with a likelihood-ratio gradient whose baseline $b_k$ comes from lifetimes against the same opponent (Section~\ref{sec:method}).} \label{fig:overview} \end{figure*} \section{Background}
\label{sec:background}

\label{sec:posg}

We formalise the underlying game as a \emph{partially observable stochastic game} (POSG) \citep{hansen_dynamic_2004} that represents general-sum games and all games in this paper, written as a tuple $\langle\mathcal I,\mathcal S,\mathcal A,\Omega,\mathcal O,\mathcal P,\mathcal R,\gamma\rangle$ of agents $\mathcal I$, states $\mathcal S$, joint actions $\mathcal A=\times_{i\in\mathcal I}\mathcal A^i$, joint observations $\Omega=\times_{i\in\mathcal I}\Omega^i$, an observation function $\mathcal O$, a transition function $\mathcal P$, rewards $\mathcal R=\times_{i\in\mathcal I}\mathcal R^i$ and a discount $\gamma\in[0,1)$. In round $t$, each agent $i$ receives an observation $o^i_t$ according to $\mathcal O$, samples an action $a^i_t\sim\pi^i(\cdot\mid o^i_t)$, and receives a reward $r^i_t=\mathcal R^i(s_t,\mathbf a_t)$, where $\mathbf a_t=(a^i_t,a^{-i}_t)$ is the joint action and $-i$ denotes the opponent.

In two-player $2\times2$ iterated matrix games, actions $\mathcal A^i=\{0,1\}$ and rewards $r^i_t=M^i[a^i_t,a^{-i}_t]$. We assign a starting state $s_\varnothing$ to the first round of each match. Thereafter, the state records the joint action as (agent $i$'s action, opponent's action), providing the state space: 
\[
\mathcal S=\{s_\varnothing,(0,0),(0,1),(1,0),(1,1)\}.
\]
A memory-one policy is conditioned on the previous joint action, with parameters $\theta$, and plays action~0 given state $s$ with probability $\pi_\theta(0\mid s)=\sigma(d_\theta(s))$, where $\sigma$ is the logistic function and $d_\theta(s)=\log\left(\frac{\pi_\theta(0\mid s)}{\pi_\theta(1\mid s)}\right)$ represents the log-odds of action~0.

To assess the performance of a fixed pair of policies, we define an agent's value as the discounted value function at the starting state and evaluate it as:
\begin{equation}
V^i(\pi^i,\pi^{-i})
=(1-\gamma)\,\mathbb E\!\left[
\sum_{t=0}^{\infty}\gamma^t r^i_t
\right].
\label{eq:exact_value}
\end{equation}

When the opponent also uses a memory-one policy, this value can be computed exactly \citep{foerster_learning_2018}. This value provides the performance measure for our
meta-training objective (Section~\ref{sec:rule_update}). This value is only provided to \method{} during training, and \method{} adapts from sampled matches without receiving this value during deployment.

\subsection{Update Rules}
\label{sec:learning}

A \emph{lifetime} is defined as a sequence of matches $k=0,1,\dots$. Each match starts in $s_\varnothing$, and the agent's policy parameters do not reset in between. Unless stated otherwise, learners update between matches using the policies from the completed match. We define agent $i$'s policy in match $k$ as $\pi^i_k$, and their value as $V^i_k=V^i(\pi^i_k,\pi^{-i}_k)$.

A \emph{naive learner} (NL) \citep{foerster_learning_2018} treats the opponent's policy as fixed when computing its own update. In our experiments, an NL playing as agent $-i$, with logits $\phi_k$ and step size $\alpha$, takes one exact gradient step after each match,
\begin{equation}
\phi_{k+1}=\phi_k+\alpha\,\nabla_{\phi}V^{-i}\big(\pi^i_k,\sigma(\phi)\big)\big|_{\phi=\phi_k},
\label{eq:nl}
\end{equation}
using agent $i$'s policy before its own update. 
\emph{Opponent shaping} instead chooses the agent's own updates to steer the opponent's learning, as LOLA and M-FOS do.

\subsection{Meta-Learning}
\label{sec:meta_learning}
Most RL algorithms are designed by hand, and researchers are the ones to decide how an agent's policy is updated. An example is the naive learner's update shown above in Equation~\eqref{eq:nl}. Meta-learning instead removes part of, or most of, the manual handcrafting aspect by learning the rule itself.

\method{} has two levels of learning. At the inner level, an inner network is represented by its parameters $\theta$. At the outer level, a meta-network is represented by meta-parameters $\eta$, and is updated to maximize the meta-objective, which, for \method{}, is an agent's return over its lifetime. In prior work, $\eta$ ranges from hyperparameters such as the discount factor \citep{xu_meta-gradient_2018} to an entire update rule \citep{oh_discovering_2020,oh_discovering_2025}, and \method{} falls more on the latter.

When meta-learning an update rule, meta-gradients through unrolled updates \citep{xu_meta-gradient_2018} can explode over long unrolls because the learning dynamics amplify small perturbations \citep{metz_understanding_2019}. In multi-agent settings, the problem worsens as the unrolls also pass through the opponent's learning. Instead, we use the likelihood-ratio gradient \citep{williams_simple_1992}, which does not differentiate through the game, the opponent's learning or the agents' parameter updates.

\section{Discovered Update Rules for Opponent Shaping}
\label{sec:method}

In iterated matrix games, after every match \method{} updates the log-odds in every state via its parameters (Figure~\ref{fig:overview}). Our key idea is to \emph{meta-learn a  rule that reads only the agent's own experience and payoff matrix}. In this section, we will first describe the architecture of the meta-network and its outputs. Then, we will analyse how the inner network takes in the meta-network output to adjust its weights accordingly. Lastly, we will talk about how  meta-training is optimised, and introduce robustness training against exploiters.

\subsection{Meta-Network}
\label{sec:rule}
The update rule is learned by training the meta-network across lifetimes. The meta-network consists of a gated recurrent unit (GRU) \citep{cho_learning_2014} and a multi-layer perceptron (MLP) \citep{albrecht_multi-agent_2024}, whose weights are denoted by meta-parameters $\eta$ (Section~\ref{sec:rule_update}). During deployment, the meta-network is frozen and the weights are fixed while \method{} uses the rule to adjust the parameters $\theta$ of its inner MLP  after
each match.

At the end of match $k$, let $\mathcal D^i_k=(s^i_t,a^i_t,a^{-i}_t,r^i_t)_{t<T}$ denote the trajectory of states, own actions, opponent actions and own rewards observed by \method{} over $T$ rounds in a match. For each state $s\in\mathcal S$, the meta-network takes in the feature vector
\[
x_{k,s}=[e_s,\ v_s,\ \hat q_s,\ p_s,\ \bar r_s]\in\mathbb R^9,
\]
where $e_s\in\mathbb R^5$ encodes the state, $v_s$ represents the fraction of rounds spent in that state,
$\hat q_s$ estimates the opponent's smoothed frequency of action~0 during match $k$,
$p_s$ is the agent's current probability of action~0,
and $\bar r_s$ is the agent's normalised mean reward in that state.

The GRU reads all five $x_{k,s}$, the mean reward and the agent's own payoff matrix after each match, and the MLP maps $x_{k,s}$ and the GRU's memory to output a scalar that represents the change to the policy's log-odds in state $s$. The rule is also applied with the two actions relabelled and the outputs averaged, so the update does not depend on what the actions are called. 

\subsection{Inner Network}
\label{sec:agent_update}

The inner network is an MLP with parameters $\theta$ that represents \method{}'s policy. With log-odds changes $y_{k,s}$ from the meta-network, we add them to the current log-odds and clip the results to obtain $d^{\mathrm{clip}}_s=\mathrm{clip}(d_{\theta_k}(s)+y_{k,s},-12,12)$. To translate $d^{\mathrm{clip}}_s$ into changes in $\theta$, we use damped Gauss-Newton steps \citep{martens_new_2020}. A small change $\delta$ in the parameter shifts the log-odds by $J\delta$, where $J$ is the Jacobian $J=\frac{\partial d_\theta}{\partial\theta}$. Hence, we require $J\delta=u$, where $u_s=d^{\mathrm{clip}}_s-d_\theta(s)$ and $J\delta=u$. We take three steps of the approximate $\delta=J^\top(JJ^\top+\Lambda)^{-1}u$, where $\Lambda$ is a damping term that makes the matrix invertible, and $J$ and $u$ are recomputed each time. The best candidate is then kept as $\theta_{k+1}$ (Appendix~\ref{sec:app-implementation}).

\subsection{Optimising Meta-Training}
\label{sec:rule_update}

\textbf{Objective:} We train the meta-network to look for meta-parameters $\eta$ whose rule produces a high value throughout a lifetime. A training lifetime has $N=100$ matches with an update after each match. We denote $V^{\mathrm{cr}}_k$ as the credited value (defined below) of the agent's policy after $k$ updates, computed with Equation~\eqref{eq:exact_value}, where $k=0$ represents the initial policy. Meta-training then maximises the weighted average
\begin{equation}
\mathcal J(\eta)=\mathbb E_{c\sim\rho,\,\tau}\Big[\textstyle\sum_{k=0}^{N}w_k\,V^{\mathrm{cr}}_k\Big],
\label{eq:objective}
\end{equation}
where the opponent $c$ is sampled from the training mixture $\rho$ (Equation~\eqref{eq:base_mixture}), $\tau$ represents a lifetime sampled with training noise, and
\[
w_k=\frac14\sum_{h\ge k}\frac{1}{h+1},\qquad h\in\{10,25,50,100\}.
\]
Thus, $\mathcal J(\eta)$ averages the mean value of the policies after $0$ to $h$ updates over the four horizons $h$, such that the rule adapts quickly and does not fall off later.

\textbf{Credit and opponents:} The credited value is the agent's value $V^{\mathrm{cr}}_k=V^i_k$ except in self-play, where both agents are credited with the mean value $\tfrac12(V^i_k+V^{-i}_k)$. The mean value is zero in IMP, so self-play does not add any gradient to the IMP rule. To train the base rule $\eta^{(0)}$, we weight lifetimes against each opponent by:
\begin{equation}
\begin{aligned}
\rho_0(\mathrm{Self}) &= 0.4,
& \rho_0(\mathrm{NL}) &= 0.4,\\
\rho_0(\mathrm{AllC}) &= 0.1,
& \rho_0(\mathrm{AllD}) &= 0.1.
\end{aligned}
\label{eq:base_mixture}
\end{equation}
Self is self-play, NL denotes a naive learner, and AllC and AllD are training-only opponents
that almost always play action~0 and action~1, respectively (cooperation and defection in IPD).

\begin{figure*}[t]   \centering   \includegraphics[width=0.9\textwidth]{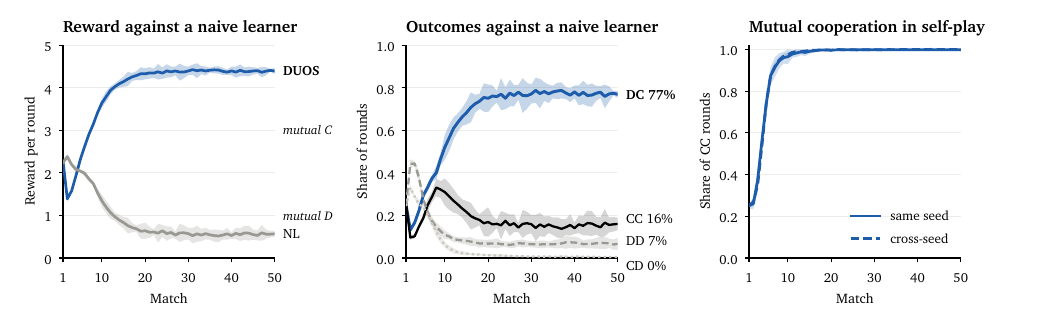}   \Description{Three line charts over the 50 matches of an IPD lifetime. Left: against a naive learner, the reward of DUOS rises to about 4.4 per round within about 20 matches, while the naive learner's reward falls to about 0.6. Centre: the share of rounds in which DUOS defects while the naive learner cooperates rises to 77 percent. Mutual cooperation settles at 16 percent and mutual defection at 7 percent. Right: in self-play, the share of mutually cooperative rounds rises from about 0.25 to 1 within about 15 matches, for copies trained with the same or different seeds.}   \caption{IPD learning trajectories over 50 matches. Left: rewards of \method{} against a naive learner (NL). Centre: joint-action frequencies against NL (\method{}'s action first, C: cooperation, D: defection). Right: mutual cooperation in self-play, within and across training seeds. Curves average five \method{} seeds. Shaded bands show 95\% $t$-intervals.}   \label{fig:ipd_behaviour} \end{figure*} \textbf{Estimator:}
Inspired by Gaussian perturbations used in evolution strategies
\citep{salimans_evolution_2017,lu_discovered_2022}, we add Gaussian
noise ($\sigma = 0.3$) to the log-odds changes during meta-training, and use the perturbed changes $\tilde y_k$ to update the inner network (Section~\ref{sec:agent_update}). We train the meta-network using a likelihood-ratio gradient estimator \citep{williams_simple_1992}, where each perturbed change contributes
\begin{equation}
\hat g_k=(G_k-b_k)\nabla_\eta
\log\bar\pi_\eta(\tilde y_k\mid\mathcal H_k).
\label{eq:meta_gradient}
\end{equation}
Here, $\bar\pi_\eta(\cdot\mid\mathcal H_k)=\mathcal N(y_k,\sigma^2I)$ is the distribution of the perturbed changes given history $\mathcal H_k=(M^i,\pi^i_j,\mathcal D^i_j)$, which consists of the agent's payoff matrix $M^i$, its policy $\pi^i_j$, and trajectory $\mathcal D^i_j$ in each match $j\le k$. The weighted value of the later policies $\theta_{k+1},\dots,\theta_N$ is $G_k=\sum_{j>k}w_j V^{\mathrm{cr}}_j$.

We sample four lifetimes from the same initial policies. For each lifetime, $b_k$ is the average of $G_k$ in the other three. This reinforces changes that beat the baseline and discourages worse ones.

We sum $\hat g_k$ over each lifetime, including both agents in self-play, and average across training lifetimes using the opponent mixture $\rho$. Adam \citep{kingma_adam_2015} is then used to update the meta-network's weights $\eta$.

\begin{algorithm}[t]
\caption{Meta-training of \method{}}
\label{alg:metatrain}
\begin{algorithmic}[1]
\Require Base mixture $\rho_0$, lifetime length $N$, PSRO rounds $L$
\State Initialise meta-parameters $\eta$
\For{$\ell=0,\dots,L$}
    \If{$\ell>0$}
        \State Train exploiter $\psi^{(\ell)}$ against $\eta$
        \State Update the meta-game $U^{(\ell)}$
        \State Compute exploiter mixture $\mu^\star$
        \State Form training mixture $\rho_\ell$
    \EndIf
    \For{each meta-training step}
        \ForAll{training lifetimes, opponent $c\sim\rho_\ell$}
            \State Initialise policies and recurrent memory
            \For{$k=0,\dots,N-1$}
                \State Evaluate $V^{\mathrm{cr}}_k$ (Eq.~\eqref{eq:exact_value})
                \State Play match $k$
                \State Sample $\tilde y_k\sim\bar\pi_\eta(\cdot\mid\mathcal H_k)$
                \State Update $\theta$ using $\tilde y_k$ (Sec.~\ref{sec:agent_update})
                \State Update the opponent (NL: Eq.~\eqref{eq:nl})
            \EndFor
            \State Evaluate $V^{\mathrm{cr}}_N$
        \EndFor
        \State Compute returns $G_k$ and baselines $b_k$
        \State Estimate $\nabla_\eta\mathcal J$ from $\hat g_k$ (Eq.~\eqref{eq:meta_gradient})
        \State Update $\eta$ with Adam
    \EndFor
    \State Store $\eta^{(\ell)}\gets\eta$
\EndFor
\State \Return $\eta^{(L)}$
\end{algorithmic}
\end{algorithm}

\subsection{Training Against Exploiters}
\label{sec:psro}

Inspired by PSRO \citep{lanctot_unified_2017} and the double oracle algorithm \citep{mcmahan_planning_2003}, each round we fix the meta-network and train an exploiter $\psi^{(\ell)}$ to maximise its own lifetime value against \method{}. The exploiter is able to observe \method{}'s current policy table and selects its policy after \method{} has updated.

We evaluate the earlier versions of the rule against all
exploiters and use the resulting meta-game matrix $U^{(\ell)}$ to compute an
exploiter mixture $\mu^\star$. This mixture minimises the highest value attained by any previous rule against it, i.e.\ each exploiter is scored by \method{}'s loss, which equals the exploiter's own gain in zero-sum IMP but not in IPD. We continue maximising $\mathcal J(\eta)$ (Equation~\eqref{eq:objective}) against a mixture $\rho_\ell$ of self-play, naive learners, the exploiter mixture and fixed strategies (Appendix~\ref{sec:app-implementation}, Supplementary Material). Each round trains the rule for 200 meta-training steps. After $L=6$ rounds, we finalise and deploy the rule (Algorithm~\ref{alg:metatrain}).

   \section{Experiments}
\label{sec:experiments}

\paragraph{Games.}
We evaluate on four iterated matrix games: Iterated Prisoner's
Dilemma (IPD), Iterated Matching Pennies (IMP), Chicken and Stag Hunt, and additionally on Coin Game. Table~\ref{tab:payoffs} shows the payoff matrices for the $2\times2$ games.
\begin{table}[H]
  \centering
  \caption{Payoff matrix for each $2\times2$ game.}
  \label{tab:payoffs}
  \small
  \setlength{\tabcolsep}{3pt}
  \renewcommand{\arraystretch}{1.15}

  \begin{minipage}[t]{0.48\linewidth}
    \centering
    \textbf{IPD}\par\smallskip
    \begin{tabular}{@{}lcc@{}}
      \toprule
                & Cooperate & Defect \\
      \midrule
      Cooperate & $(3,3)$   & $(0,5)$ \\
      Defect    & $(5,0)$   & $(1,1)$ \\
      \bottomrule
    \end{tabular}
  \end{minipage}\hfill
  \begin{minipage}[t]{0.48\linewidth}
    \centering
    \textbf{IMP}\par\smallskip
    \begin{tabular}{@{}lcc@{}}
      \toprule
            & Heads    & Tails \\
      \midrule
      Heads & $(1,-1)$ & $(-1,1)$ \\
      Tails & $(-1,1)$ & $(1,-1)$ \\
      \bottomrule
    \end{tabular}
  \end{minipage}

  \par\medskip

  \begin{minipage}[t]{0.48\linewidth}
    \centering
    \textbf{Chicken}\par\smallskip
    \begin{tabular}{@{}lcc@{}}
      \toprule
               & Swerve  & Straight \\
      \midrule
      Swerve   & $(3,3)$ & $(1,4)$ \\
      Straight & $(4,1)$ & $(0,0)$ \\
      \bottomrule
    \end{tabular}
  \end{minipage}\hfill
  \begin{minipage}[t]{0.48\linewidth}
    \centering
    \textbf{Stag Hunt}\par\smallskip
    \begin{tabular}{@{}lcc@{}}
      \toprule
           & Stag    & Hare \\
      \midrule
      Stag & $(4,4)$ & $(0,3)$ \\
      Hare & $(3,0)$ & $(2,2)$ \\
      \bottomrule
    \end{tabular}
  \end{minipage}
\end{table}

\paragraph{Coin Game.}
Coin Game \citep{lerer_maintaining_2018} is a sequential social dilemma game that extends IPD to a grid world, and the rules are given in Appendix~\ref{sec:app-coin-rules} (Supplementary Material). Because Coin Game has too many states for per-state log-odds changes, we train a variant of \method{} for it: after each episode, its rule instead outputs a multiplier $\alpha_c\in(-1,1)$ for each pickup scenario $c$ (which coin each player last collected), and the inner network takes one Adam step along $\sum_t\alpha_{c_t}\hat A_t\nabla_\theta\log\pi_\theta(a_t\mid o_t,c_t)$, where $\hat A_t$ is an advantage estimate. Adam's second moment uses the gradient with $\alpha\equiv1$, so the step scales with $\alpha$. Hence, where $\alpha\equiv1$, it is exactly like REINFORCE \citep{williams_simple_1992} (Table~\ref{tab:hp-DUOS-coin}, Supplementary Material).

\paragraph{Deployment.}
At deployment, \method{}'s meta-network is frozen and no
training noise is added. In each match, \method{} plays against
its opponent for 20 rounds using its current policy and updates its inner network for the next match. The inner-network
parameters and recurrent memory persist across matches and reset at the start of each lifetime. Unlike meta-training, tournaments score undiscounted 20-round matches over 50-match lifetimes and use a naive learner that learns faster than any seen in training. 

\begin{figure*}[t]
  \centering
  \includegraphics[width=0.9\textwidth]{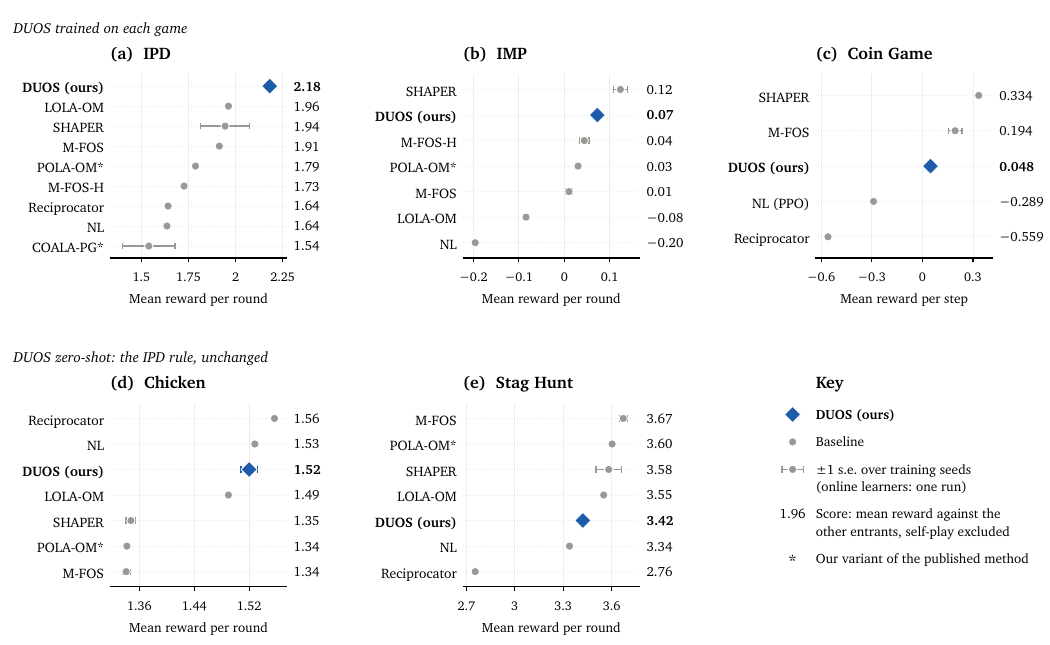}
  \Description{Five dot plots of tournament scores, one per game, with the methods sorted by score and DUOS highlighted. Error bars show one standard error over training seeds. Top row, DUOS trained in each game: it ranks first in IPD (2.18 per round), second in IMP (0.07, after SHAPER) and third in Coin Game (0.048 per step, after SHAPER and M-FOS). Bottom row, DUOS playing its IPD rule zero-shot: third in Chicken (1.52) and fifth in Stag Hunt (3.42).}
  \caption{Tournament scores after 50 matches (reward per round) or 16 episodes for Coin Game (reward per step). Self-play is excluded. Top row: \method{} trained on each game. Bottom row: \method{} trained on IPD playing zero-shot.}
  \label{fig:tournament_scores}
\end{figure*} \paragraph{Baselines.}

We run rigorous reproduction tests before using the baselines for comparison against our method. Each baseline uses its authors' settings and is checked against their results (where comparable) before entering the tournament (Appendix~\ref{sec:app-baselines}, Supplementary Material). We flag SHAPER \citep{khan_scaling_2024}, which falls slightly short in IPD. M-FOS passes and performs better in Coin Game than reported. Reciprocator passes with its released code but not with the paper's published formula, so we use the released code, and its Coin Game check is also flagged. Advantage Alignment \citep{duque_advantage_2025} and LOQA \citep{aghajohari_loqa_2024} did not pass and are excluded. All hyperparameters and protocols are in Appendix~\ref{sec:app-hyperparameters} (Supplementary Material).

\paragraph{Evaluation.}
One tournament is held per game. \method{}, M-FOS
\citep{lu_model-free_2022}, SHAPER \citep{khan_scaling_2024} and NL participate in all five games. LOLA-OM \citep{foerster_learning_2018} and POLA-OM* participate in the four matrix games, where POLA-OM* builds on Outer-POLA \citep{zhao_proximal_2022}, which computes POLA's proximal update from exact values for tabular policies. Like POLA's opponent-modelling variant (POLA-OM), POLA-OM* replaces the opponent's policy with a model fitted to its observed actions.
IPD and IMP additionally include the history-based M-FOS variant
(M-FOS-H) \citep{lu_model-free_2022}.
COALA-PG*, our pure-shaping reconstruction of COALA-PG
\citep{meulemans_multi-agent_2025}, is included in IPD.
Reciprocator \citep{zhou_reciprocal_2024} is included in IPD,
Chicken, Stag Hunt and Coin Game.

In the matrix games, every pairing plays 128 lifetimes $\times$ 50 matches $\times$ 20 rounds. We evaluate five training seeds for \method{}, M-FOS, M-FOS-H and COALA-PG*, three for SHAPER, and none for LOLA-OM, POLA-OM*, Reciprocator and NL because they learn online and do not require training.

Information access differs between methods: \method{} uses only its own payoff matrix and observed match trajectories, whereas M-FOS, NL and Reciprocator receive the opponent's policy, with memory-one approximations for recurrent opponents. Reciprocator also uses the opponent's payoff matrix and 2{,}048 unscored practice matches of 20 rounds per update against the opponent. \mbox{M-FOS-H} reads its own policy and the 20 joint actions of the last match through a memory that resets after each match. LOLA-OM and POLA-OM* use observed actions and both payoff matrices and assume an NL update. SHAPER and COALA-PG* only use their own history.

\begin{table*}[t]
  \centering
  \small
  \caption{Normalised tournament scores. Each game's scores are normalised in accordance with the scoring criteria (1 = best, 0 = worst). Matrix mean: four $2 \times 2$ matrix games. Five-game mean: all five games for the methods that play all five (95\% seed-bootstrap intervals). Rank: by five-game mean. NL in Coin Game is a PPO learner \cite{schulman_proximal_2017}. Bold indicates the best result in each column.}
  \label{tab:normalised_scores}\label{tab:normalised}
  \begin{tabular}{lrrrrrrlr}
    \toprule
    Method & IPD & IMP & Chicken & Stag Hunt & Coin & Matrix mean & Five-game mean & Rank \\
    \midrule
    \method{} (ours) & \textbf{1.000} & 0.841 & 0.829 & 0.727 & 0.680 & \textbf{0.849} & \textbf{0.815} {\scriptsize[0.78, 0.85]} & \textbf{1} \\
    SHAPER & 0.631 & \textbf{1.000} & 0.032 & 0.902 & \textbf{1.000} & 0.641 & 0.713 {\scriptsize[0.64, 0.78]} & 2 \\
    M-FOS & 0.583 & 0.646 & 0.000 & \textbf{1.000} & 0.844 & 0.557 & 0.614 {\scriptsize[0.57, 0.66]} & 3 \\
    NL & 0.150 & 0.000 & 0.868 & 0.637 & 0.302 & 0.414 & 0.391 {\scriptsize[0.35, 0.43]} & 4 \\
    \midrule
    LOLA-OM & 0.659 & 0.350 & 0.690 & 0.868 & -- & 0.642 & -- & -- \\
    POLA-OM* & 0.386 & 0.708 & 0.005 & 0.925 & -- & 0.506 & -- & -- \\
    M-FOS-H & 0.292 & 0.751 & -- & -- & -- & -- & -- & -- \\
    Reciprocator & 0.159 & -- & \textbf{1.000} & 0.000 & 0.000 & -- & -- & -- \\
    COALA-PG* & 0.000 & -- & -- & -- & -- & -- & -- & -- \\
    \bottomrule
  \end{tabular}
\end{table*} \paragraph{Scoring.}
A method's score is defined as its undiscounted mean reward per round (per step for Coin Game), averaged equally over its opponents, excluding self-play.
To compare across varying payoff scales, we normalise the score of each game:
\[
n_g(a)=
\frac{s_g(a)-\min_{b\in\mathcal E_g}s_g(b)}
     {\max_{b\in\mathcal E_g}s_g(b)-\min_{b\in\mathcal E_g}s_g(b)},
\]
where $s_g(a)$ is method $a$'s score and $\mathcal E_g$ is the set of entrants in game $g$. The normalised mean score averages $n_g(a)$ over all games equally.  In IMP, $s_g$ weights both player roles equally.
\paragraph{Zero-shot transfer.}
To observe effects of generalisation, \method{} plays Chicken and Stag Hunt zero-shot with its IPD rule as well as in self-play.

\section{Results}

\subsection{Head-to-Head}
\label{sec:results} Figure~\ref{fig:tournament_scores} and Table~\ref{tab:normalised_scores} compare the tournament scores across all five games. The full matchup results are given in Supplementary Material (Appendix~\ref{sec:app-tournaments}).
\method{} ranks \textit{1st} by normalised mean score, both over the four $2 \times 2$ matrix games
and over all five games, where it plays zero-shot in Chicken and Stag Hunt, trained only in IPD. We only rank methods that have participated in all of
the five games. Individually,
\method{} ranks \textit{1st} in IPD, \textit{2nd} in IMP, \textit{3rd} in Chicken, \textit{5th} in
Stag Hunt and \textit{3rd} in Coin Game.

\textbf{Opponent shaping:} Figure~\ref{fig:ipd_behaviour} shows that in IPD, both players are initially cooperating in about half of the rounds. By the end, NL has cooperated in 92.9\% of the rounds and \method{} only in 16.4\%. As such, \method{}'s mean reward per round during its final rounds has doubled compared to its initial rounds, and vice versa for NL. Separately, we replay each lifetime with \method{}'s inner network frozen at its final policy from the first match, and show that NL cooperates in only 14.1\% of the rounds (Table~\ref{tab:app-frozen}, Supplementary Material). Therefore, NL's cooperation is produced by how \method{}'s policy changes over time, not by its final strategy alone, which is consistent with opponent shaping behaviour.

\textbf{IPD:} \method{} obtains the highest mean score in IPD despite losing most head-to-head matchups (6 out of 8, Table~\ref{tab:app-h2h}, Supplementary Material). Head-to-head matchups do not fully translate to final rankings in general-sum games because the scores count the reward each method obtains, not its margin over the opponent. For example, winning $2.10$ to $1.30$ scores less than mutual cooperation at $3.0$ each. This reflects the meta-objective (Equation~\eqref{eq:objective}), which maximises expected return rather than head-to-head margins. This finding mirrors Axelrod's tournaments, where tit-for-tat won on total score without outscoring any opponent \citep{axelrod_evolution_2006}.

\begin{table*}[t]
\centering
\small
\caption{Normalised self-play scores: each method's mean own reward against a same-seed copy of itself per round, averaged over a lifetime of 50 matches (per step over 16 episodes in Coin Game). Each game's scores are normalised in accordance with the scoring criteria. IMP is omitted because it is zero-sum. $^{\ddagger}$Outside the tournament (two learners of each run play each other).}
\label{tab:selfplay_scores}
\begin{tabular}{lrrrrrlr}
\toprule
Method & IPD & Chicken & Stag Hunt & Coin & Matrix mean & Four-game mean & Rank \\
\midrule
\textbf{\method{} (ours)} & \textbf{1.000} & \textbf{1.000} & 0.883 & \textbf{1.000} & \textbf{0.961} & \textbf{0.971} {\scriptsize[0.969, 0.973]} & \textbf{1} \\
Reciprocator & 0.621 & 0.804 & 0.000 & 0.954 & 0.475 & 0.595 {\scriptsize[0.590, 0.601]} & 2 \\
NL & 0.031 & 0.848 & 0.101 & 0.900 & 0.327 & 0.470 {\scriptsize[0.465, 0.476]} & 3 \\
M-FOS & 0.005 & 0.073 & 0.970 & 0.774 & 0.349 & 0.456 {\scriptsize[0.396, 0.499]} & 4 \\
SHAPER & 0.327 & 0.000 & \textbf{1.000} & 0.000 & 0.442 & 0.332 {\scriptsize[0.274, 0.364]} & 5 \\
\midrule
LOLA-OM & 0.428 & 0.516 & 0.554 & -- & 0.499 & -- & -- \\
POLA-OM* & 0.058 & 0.079 & 0.782 & -- & 0.306 & -- & -- \\
COALA-PG* & 0.091 & -- & -- & -- & -- & -- & -- \\
M-FOS-H & 0.000 & -- & -- & -- & -- & -- & -- \\
\midrule
M-FOS (meta-self-play)$^{\ddagger}$ & 1.028 & -- & -- & 1.248 & -- & -- & -- \\
\bottomrule
\end{tabular}
\end{table*}

\begin{figure*}[t]
\centering
\includegraphics[width=0.9\textwidth]{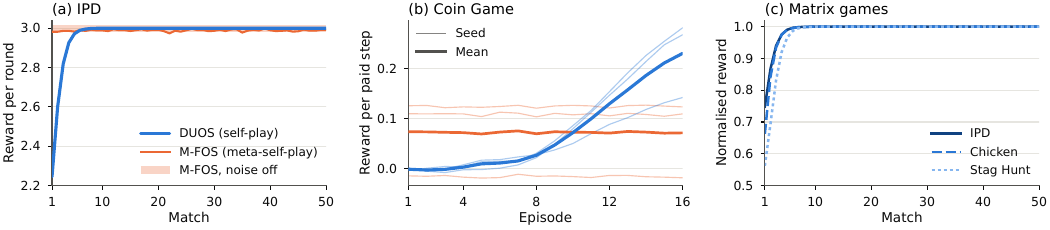}
\Description{Three line charts. (a) IPD: the self-play reward of DUOS rises from 2.25 to 3 within about ten matches. M-FOS (meta-self-play) stays just below 3 from the first match, and its noise-off version is at 3 throughout. (b) Coin Game: DUOS rises from 0 to about 0.23 over 16 episodes while M-FOS stays flat near 0.07. Thin lines show the individual seeds. (c) DUOS (self-play) reaches the mutual-cooperation payoff within about ten matches in IPD, Chicken and Stag Hunt.}
\caption{Learning curves of the variants trained for self-play. (a)~IPD: mean over 5 seeds (s.e.\ $\le 0.009$). (b)~Coin Game (3 seeds), reward per step. (c)~\method{} (self-play) in the three matrix games with normalised reward: reward per round divided by the game's mutual-cooperation payoff (3, 3 and 4), s.e.\ $\le 0.007$.}
\label{fig:selfplay_curves}
\end{figure*}

\textbf{IMP:} \method{} wins against every opponent in IMP, and is the only method that SHAPER does not beat (Table~\ref{tab:imp_matrix}, Supplementary Material). This can be partly credited to PSRO training, which lowers \method{}'s exploitability from $0.376$ to $0.144$ and reduces how much M-FOS and SHAPER, which specialise in exploiting learners, gain against it (Table~\ref{tab:app-psro}, Supplementary Material).

\textbf{Chicken:} Playing Chicken zero-shot, \method{} ranks \textit{3rd} with $1.52$ per round, and \textit{1st} without SHAPER participating ($1.81$, Table~\ref{tab:shaper-original-vs-extended}, Supplementary Material). SHAPER and M-FOS play Straight in Chicken and \method{} does not yield to them, so \method{} ends up mostly crashing, which lowers its score. This might be due to SHAPER and M-FOS training only against naive learners, and thus learning mainly to exploit. Consistent with this, SHAPER and M-FOS have the two lowest Chicken self-play scores (Table~\ref{tab:selfplay_scores}). The Chicken result suggests that learned rules transfer across different settings, as also observed in single-agent reinforcement learning \citep{oh_discovering_2025}, although a rule trained on Chicken itself still scores higher ($2.105$ vs.\ $1.810$, Table~\ref{tab:app-transfer}, Supplementary Material).

\vskip 0pt plus 3pt\relax\textbf{Stag Hunt:} \method{} plays Stag Hunt zero-shot and ranks \textit{5th} in the tournament with a score of $3.42$, only $0.25$ behind M-FOS in \textit{1st}. Yet, \method{} wins every head-to-head matchup here, the opposite of its IPD result. In Stag Hunt, a player out-earns its opponent only by playing Hare (3) against Stag (0), giving up mutual Stag (4), on which the other methods coordinate. Most of the gap is the cost of transferring from IPD: in the tournament without SHAPER, when trained in Stag Hunt itself rather than in IPD, \method{} scores $0.25$ more and ranks \textit{2nd} instead of \textit{4th} (Table~\ref{tab:app-transfer}, Supplementary Material).

\textbf{Coin Game:} \method{} ranks \textit{3rd} of five with $0.048$ per step (Table~\ref{tab:coin-tournament}, Supplementary Material). Head to head, \method{} out-earns NL and Reciprocator but loses to M-FOS and SHAPER. \method{} learns a rule that is akin to REINFORCE \citep{williams_simple_1992}: its multiplier $\alpha_c$ is always around $1.00$, and thus \method{} learns like a plain ($\alpha = 1$) learner rather than shaping its opponent. This is likely a limit of the rule form, which can only rescale the REINFORCE step by a multiplier in $(-1,1)$.

\subsection{Self-Play}
A widely adopted algorithm will often face copies of itself. M-FOS reaches cooperation in self-play through a meta-self-play curriculum \citep{lu_model-free_2022}. \method{} is trained against a population that includes 40\% self-play. In IPD, \method{} reaches mutual cooperation (Figure~\ref{fig:ipd_behaviour}, right), with both players cooperating in 93.4\% of rounds, eventually reaching 99.8\% during the final rounds. We pair different training seeds with each other, and also observe similar trajectories, showing that cooperation extends beyond exact same copies. \method{} therefore behaves differently against NL (Figure~\ref{fig:ipd_behaviour}) than against another copy of itself: its changing policy makes NL more cooperative, which it then exploits, but it reciprocates mutual cooperation in self-play. These trajectories are consistent with opponent shaping.

\textbf{Tournaments:} \method{} clearly outperforms all baselines in self-play score (Table~\ref{tab:selfplay_scores}) and ranks \textit{1st} in 3 out of 4 games against methods tested in all four (deployed zero-shot in Chicken and Stag Hunt). Outside of the tournaments, M-FOS (meta-self-play) scores higher than \method{}, which we suspect stems from the different training mixtures. Hence, we train an exploratory self-play variant of \method{} with one rule shared by both copies and compare it with M-FOS (meta-self-play) in Figure~\ref{fig:selfplay_curves} (Appendix~\ref{sec:app-results}). 

\textbf{\method{} (self-play) vs. M-FOS (meta-self-play):} The two methods are trained differently. \method{} (self-play) is trained only in self-play from the start with no naive learners, fixed strategies or PSRO. M-FOS (meta-self-play) instead follows its meta-self-play curriculum, where two separately initialised learners play each other with increasing probability, and otherwise play against naive learners. \method{} (self-play) is trained on 256{,}000 self-play lifetimes in IPD and 243{,}200 in Coin Game, whereas M-FOS (meta-self-play) is trained on around 2.8 million and around 900 lifetimes, respectively. In IPD, M-FOS (meta-self-play) nears mutual cooperation from the first match, and is fully cooperative with sampling noise off. \method{} (self-play) starts from a fresh random policy (with a trained meta-network) and fully converges by match 9--10. In Coin Game, the reward of M-FOS (meta-self-play) stays roughly constant, while \method{} (self-play) stays near zero for the first few episodes and then increases, overtaking M-FOS (meta-self-play) after episode 10. Over the lifetime, both methods are level (0.075 vs. 0.073 per step). Trained in each game, \method{} (self-play) also converges to mutual cooperation in Chicken and Stag Hunt, akin to its behaviour in IPD.

\section{Conclusion}
In single-agent reinforcement learning, discovered algorithms \citep{goldie_how_2025,goldie_can_2024,jackson_discovering_2024} have achieved state-of-the-art results \citep{oh_discovering_2025}. We introduce \method{}, to our knowledge the first method to discover an update rule for an agent in general-sum games. Without reading the opponent's policies or parameters, \method{} ranks \textit{1st} in normalised mean score across all games, and \textit{1st} for self-play among tournament entrants. In IPD, it exploits a naive learner while still cooperating with a copy of itself, consistent with opponent shaping. The IPD-trained rule also transfers zero-shot to two other $2\times2$ matrix games, Chicken and Stag Hunt, with competitive scores, as discovered rules also do in single-agent settings \citep{oh_discovering_2025}. Meta-training against exploiters (PSRO) makes the rule harder to exploit, lowering its exploitability in IMP by more than half. In Coin Game, \method{}'s learned rule reduces to REINFORCE ($\alpha \approx 1$) and ranks \textit{3rd} of 5 entrants. In self-play, \method{} (self-play) achieves full mutual cooperation in IPD, Chicken and Stag Hunt within 12 matches from a fresh policy, and eventually obtains more reward per step than M-FOS (meta-self-play) after 10 episodes in Coin Game. Our work has limitations. Each rule is trained per game, and in Chicken, a rule trained on Chicken specifically still scores higher than the IPD rule played zero-shot. In Coin Game, shaping is likely limited by the rule form, which can only rescale the REINFORCE step. Future work could investigate meta-training one rule across several games, as DiscoRL's discovered rule improved with more compute and more diverse training environments \citep{oh_discovering_2025}. We could also investigate different rule forms and training settings that shape opponents and generalise to different sequential social dilemma games \citep{leibo_multi-agent_2017} such as Harvest \citep{perolat_multi-agent_2017,hughes_inequity_2018} and Cleanup \citep{hughes_inequity_2018}.

\bibliographystyle{ACM-Reference-Format} 
\bibliography{references} \clearpage \onecolumn \raggedbottom %
\setcounter{table}{0}\renewcommand{\thetable}{S\arabic{table}}\renewcommand{\theHtable}{S\arabic{table}}
\setcounter{figure}{0}\renewcommand{\thefigure}{S\arabic{figure}}\renewcommand{\theHfigure}{S\arabic{figure}}
\section*{Supplementary Material} \label{sec:supplementary} \appendix \section{Coin Game rules}\label{sec:app-coin-rules}
We use the two-coin version of Coin Game \citep{lerer_maintaining_2018} implemented in pax by SHAPER \citep{khan_scaling_2024}. There are two players, Red and Blue, playing on a $3\times3$ grid whose edges wrap around (a torus). The grid holds one red and one blue coin, owned by Red and Blue respectively. The players and coins can share the same cells. At each step, both players move simultaneously and choose one of the five actions: right, left, up, down or stay. A player collects a coin by moving to or staying on the cell the coin is occupying. Collecting any coin gives the collector $+1$. If the coin belongs to the other player, the other player also receives $-2$. If both players collect the same coin, both receive $+1$ and its owner also receives $-2$ (net $-1$). A collected coin reappears on a uniformly random cell. An episode starts with the four positions drawn independently and uniformly over nine cells. The episode lasts 16 steps, of which 15 are paid: the 16th step applies the moves and resets the board, giving both players zero reward. Throughout the paper, a Coin Game reward per step is an episode's total reward divided by its 15 paid steps.
 \section{Tournament results}\label{sec:app-tournaments} Tables~\ref{tab:ipd_matrix}, \ref{tab:imp_matrix}, \ref{tab:chicken-shaper}, \ref{tab:stag-shaper} and~\ref{tab:coin-tournament} give every head-to-head result of the tournaments in Table~\ref{tab:normalised}. SHAPER was trained for Chicken and Stag Hunt and added to those fields after the first scores had been computed. Table~\ref{tab:normalised-original} gives the normalised scores without it, and Table~\ref{tab:shaper-original-vs-extended} compares the fields without and with it. Table~\ref{tab:psro_control_scores} replaces \method{} with its pre-PSRO base rule in the same fields, with the same random draws. \begin{table}[H] \centering
  \caption{IPD tournament after 50 matches. Each entry is the row method's mean reward per round against the column method. Self-play is excluded from the mean score. \method{}'s diagonal entry pools same-seed and cross-seed pairings of its training seeds. \method{}'s row is in bold.}
  \label{tab:ipd_matrix}
  \resizebox{ 0.85\columnwidth}{!}{%
    \begin{tabular}{lccccccccccc}
      \toprule
      Row vs.\ column & M-FOS & M-FOS-H & SHAPER & COALA-PG* & LOLA-OM & POLA-OM* & NL & Reciprocator & \method{} & Mean score & Rank \\
      \midrule
      M-FOS        & --   & 1.05 & 1.42 & 1.25 & 1.66 & 1.43 & 3.88 & 2.61 & 2.00 & 1.91 & 4 \\
      M-FOS-H      & 1.04 & --   & 1.38 & 1.26 & 1.36 & 1.11 & 3.13 & 2.45 & 2.09 & 1.73 & 6 \\
      SHAPER       & 1.35 & 1.29 & --   & 1.34 & 1.92 & 1.61 & 3.53 & 2.39 & 2.13 & 1.94 & 3 \\
      COALA-PG*    & 1.18 & 1.13 & 1.36 & --   & 1.46 & 1.36 & 2.02 & 2.18 & 1.62 & 1.54 & 9 \\
      LOLA-OM      & 1.53 & 1.31 & 1.77 & 1.37 & --   & 1.34 & 2.90 & 2.72 & 2.76 & 1.96 & 2 \\
      POLA-OM*     & 1.36 & 1.09 & 1.59 & 1.36 & 1.39 & --   & 3.75 & 2.28 & 1.48 & 1.79 & 5 \\
      NL           & 1.08 & 1.83 & 1.57 & 1.93 & 2.60 & 1.26 & --   & 1.93 & 0.90 & 1.64 & 8 \\
      Reciprocator & 1.55 & 1.58 & 1.83 & 1.55 & 2.47 & 1.24 & 1.08 & --   & 1.82 & 1.64 & 7 \\
      \textbf{\method{} (ours)} & \textbf{1.73} & \textbf{1.83} & \textbf{1.93} & \textbf{1.61} & \textbf{2.58} & \textbf{1.28} & \textbf{3.96} & \textbf{2.53} & \textbf{2.93} & \textbf{2.18} & \textbf{1} \\
      \bottomrule
    \end{tabular}%
  }
\end{table}

\begin{table}[H] \centering
  \caption{IMP tournament after 50 matches. Each entry is the row method's mean reward per round against the column method, averaged over the two seats (so the matrix is antisymmetric). Self-play is excluded from the mean score. \method{}'s diagonal entry pools same-seed and cross-seed pairings of its training seeds. \method{}'s row is in bold.}
  \label{tab:imp_matrix}
  \resizebox{ 0.85\columnwidth}{!}{%
    \begin{tabular}{lccccccccc}
      \toprule
      Row vs.\ column & M-FOS & M-FOS-H & SHAPER & LOLA-OM & POLA-OM* & NL & \method{} & Mean score & Rank \\
      \midrule
      M-FOS   & --    & $-$0.04 & $-$0.35 &  0.16 & $-$0.12 &  0.47 & $-$0.06 &  0.01 & 5 \\
      M-FOS-H &  0.04 & --    & $-$0.15 &  0.04 & $-$0.05 &  0.40 & $-$0.01 &  0.04 & 3 \\
      SHAPER  &  0.35 &  0.15 & --    &  0.11 &  0.01 &  0.13 & $-$0.01 &  0.12 & 1 \\
      LOLA-OM & $-$0.16 & $-$0.04 & $-$0.11 & --    & $-$0.02 & $-$0.02 & $-$0.16 & $-$0.08 & 6 \\
      POLA-OM* &  0.12 &  0.05 & $-$0.01 &  0.02 & --    &  0.00 &  0.00 &  0.03 & 4 \\
      NL      & $-$0.47 & $-$0.40 & $-$0.13 &  0.02 &  0.00 & --    & $-$0.19 & $-$0.20 & 7 \\
      \textbf{\method{} (ours)} & \textbf{0.06} & \textbf{0.01} & \textbf{0.01} & \textbf{0.16} & \textbf{0.00} & \textbf{0.19} & \textbf{0.00} & \textbf{0.07} & \textbf{2} \\
      \bottomrule
    \end{tabular}%
  }
\end{table}

\begin{table}[H]
\centering
\small
\caption{Chicken tournament after 50 matches, with SHAPER: the row method's mean reward per round against each column method. \method{} plays its IPD rule zero-shot. Its self-play (diagonal) entry pools same-seed and cross-seed pairings of its training seeds and is excluded from the score.}
\label{tab:chicken-shaper}
\begin{tabular}{lrrrrrrrrr}
\toprule
Row vs.\ column & M-FOS & SHAPER & LOLA-OM & POLA-OM* & NL & Reciprocator & \method{} & Mean score & Rank \\
\midrule
M-FOS & -- & 0.06 & 0.38 & 0.15 & 3.91 & 3.23 & 0.30 & 1.34 & 7 \\
SHAPER & 0.26 & -- & 0.23 & 0.11 & 3.92 & 3.29 & 0.27 & 1.35 & 5 \\
LOLA-OM & 0.11 & 0.06 & -- & 0.40 & 3.18 & 2.65 & 2.54 & 1.49 & 4 \\
POLA-OM* & 0.09 & 0.03 & 0.51 & -- & 3.87 & 2.83 & 0.73 & 1.34 & 6 \\
NL & 0.98 & 0.98 & 2.34 & 1.04 & -- & 2.08 & 1.74 & 1.53 & 2 \\
Reciprocator & 0.82 & 0.82 & 2.36 & 0.81 & 2.69 & -- & 1.83 & 1.56 & 1 \\
\textbf{\method{} (ours)} & 0.11 & 0.07 & 2.44 & 0.51 & 3.51 & 2.48 & 2.92 & 1.52 & 3 \\
\bottomrule
\end{tabular}
\end{table}

\begin{table}[H]
\centering
\small
\caption{Stag Hunt tournament after 50 matches, with SHAPER: the row method's mean reward per round against each column method. \method{} plays its IPD rule zero-shot. Its self-play (diagonal) entry pools same-seed and cross-seed pairings of its training seeds and is excluded from the score.}
\label{tab:stag-shaper}
\begin{tabular}{lrrrrrrrrr}
\toprule
Row vs.\ column & M-FOS & SHAPER & LOLA-OM & POLA-OM* & NL & Reciprocator & \method{} & Mean score & Rank \\
\midrule
M-FOS & -- & 3.97 & 3.77 & 3.89 & 3.93 & 2.91 & 3.57 & 3.67 & 1 \\
SHAPER & 3.97 & -- & 3.63 & 3.91 & 3.82 & 2.82 & 3.34 & 3.58 & 3 \\
LOLA-OM & 3.83 & 3.68 & -- & 3.51 & 3.74 & 3.07 & 3.47 & 3.55 & 4 \\
POLA-OM* & 3.92 & 3.93 & 3.53 & -- & 3.76 & 2.91 & 3.57 & 3.60 & 2 \\
NL & 3.94 & 3.85 & 3.73 & 3.75 & -- & 2.26 & 2.49 & 3.34 & 6 \\
Reciprocator & 3.33 & 3.12 & 3.15 & 2.95 & 1.92 & -- & 2.06 & 2.76 & 7 \\
\textbf{\method{} (ours)} & 3.71 & 3.46 & 3.56 & 3.63 & 3.36 & 2.81 & 3.79 & 3.42 & 5 \\
\bottomrule
\end{tabular}
\end{table}

\begin{table}[H]
\centering
\small
\caption{Coin Game tournament after 16 episodes: the row method's mean reward per step against each column method. \method{}'s self-play (diagonal) entry pools same-seed and cross-seed pairings of its training seeds and is excluded from the score. NL is M-FOS's PPO naive learner. M-FOS (stronger than in its paper) and Reciprocator are flagged (Appendix~\ref{sec:app-baselines}).}
\label{tab:coin-tournament}
\begin{tabular}{lrrrrrrr}
\toprule
Row vs.\ column & M-FOS & SHAPER & NL (PPO) & Reciprocator & \method{} & Mean score & Rank \\
\midrule
M-FOS & -- & $-$0.298 & 0.411 & 0.567 & 0.097 & 0.194 & 2 \\
SHAPER & 0.032 & -- & 0.534 & 0.643 & 0.127 & 0.334 & 1 \\
NL (PPO) & $-$0.447 & $-$0.754 & -- & 0.259 & $-$0.216 & $-$0.289 & 4 \\
Reciprocator & $-$0.618 & $-$0.865 & $-$0.272 & -- & $-$0.482 & $-$0.559 & 5 \\
\textbf{\method{} (ours)} & $-$0.119 & $-$0.373 & 0.230 & 0.454 & 0.014 & 0.048 & 3 \\
\bottomrule
\end{tabular}
\end{table}

\begin{table}[H]
\centering
\small
\caption{Normalised tournament scores in which SHAPER plays only IPD, IMP and Coin Game. Columns are as in Table~\ref{tab:normalised}, without the rank column. SHAPER was only added to Chicken and Stag Hunt after these scores were computed.}
\label{tab:normalised-original}
\begin{tabular}{lrrrrrrl}
\toprule
Method & IPD & IMP & Chicken & Stag Hunt & Coin & Matrix mean & Five-game mean \\
\midrule
\textbf{\method{} (ours)} & \textbf{1.000} & 0.841 & \textbf{1.000} & 0.787 & 0.680 & \textbf{0.907} & \textbf{0.862} {\scriptsize[0.84, 0.88]} \\
M-FOS & 0.583 & 0.646 & 0.000 & \textbf{1.000} & 0.844 & 0.557 & 0.614 {\scriptsize[0.57, 0.66]} \\
NL & 0.150 & 0.000 & 0.196 & 0.596 & 0.302 & 0.236 & 0.249 {\scriptsize[0.21, 0.29]} \\
\midrule
LOLA-OM & 0.659 & 0.350 & 0.842 & 0.905 & -- & 0.689 & -- \\
POLA-OM* & 0.386 & 0.708 & 0.040 & 0.920 & -- & 0.514 & -- \\
SHAPER & 0.631 & \textbf{1.000} & -- & -- & \textbf{1.000} & -- & -- \\
M-FOS-H & 0.292 & 0.751 & -- & -- & -- & -- & -- \\
Reciprocator & 0.159 & -- & 0.502 & 0.000 & 0.000 & -- & -- \\
COALA-PG* & 0.000 & -- & -- & -- & -- & -- & -- \\
\bottomrule
\end{tabular}
\end{table}

\begin{table}[H]
\centering
\small
\caption{Chicken and Stag Hunt scores (rank) after 50 matches, without SHAPER (original field) and with SHAPER.}
\label{tab:shaper-original-vs-extended}
\begin{tabular}{lllll}
\toprule
 & \multicolumn{2}{c}{Chicken} & \multicolumn{2}{c}{Stag Hunt} \\
\cmidrule(lr){2-3}\cmidrule(lr){4-5}
Method & without & with & without & with \\
\midrule
M-FOS & 1.60 (6) & 1.34 (7) & 3.61 (1) & 3.67 (1) \\
SHAPER & -- & 1.35 (5) & -- & 3.58 (3) \\
LOLA-OM & 1.78 (2) & 1.49 (4) & 3.52 (3) & 3.55 (4) \\
POLA-OM* & 1.60 (5) & 1.34 (6) & 3.54 (2) & 3.60 (2) \\
NL & 1.64 (4) & 1.53 (2) & 3.24 (5) & 3.34 (6) \\
Reciprocator & 1.70 (3) & 1.56 (1) & 2.68 (6) & 2.76 (7) \\
\textbf{\method{} (ours)} & 1.81 (1) & 1.52 (3) & 3.42 (4) & 3.42 (5) \\
\bottomrule
\end{tabular}
\end{table}

\begin{table}[H] \centering
  \caption{Mean score (rank in its own field) of \method{} with and without PSRO after $K$ matches, and their difference (with PSRO minus pre-PSRO) paired by seed with a 95\% $t$-interval. Bold: the interval excludes 0. Chicken and Stag Hunt: fields with SHAPER.}
  \label{tab:psro_control_scores}
    \begin{tabular}{lcccc}
      \toprule
      Game & $K$ & With PSRO & Pre-PSRO & Difference \\
      \midrule
      IPD & 50 & 2.182 (1) & 2.162 (1) & +0.020 {\scriptsize[$-$0.151, +0.191]} \\
       & 10 & 1.945 (1) & 1.922 (3) & +0.023 {\scriptsize[$-$0.024, +0.069]} \\
      \midrule
      IMP & 50 & 0.073 (2) & 0.029 (4) & \textbf{+0.043 {\scriptsize[+0.022, +0.065]}} \\
       & 10 & 0.054 (2) & 0.031 (2) & \textbf{+0.023 {\scriptsize[+0.009, +0.037]}} \\
      \midrule
      Chicken & 50 & 1.520 (3) & 1.531 (2) & $-$0.011 {\scriptsize[$-$0.055, +0.034]} \\
       & 10 & 1.362 (2) & 1.367 (2) & $-$0.005 {\scriptsize[$-$0.032, +0.023]} \\
      \midrule
      Stag Hunt & 50 & 3.422 (5) & 3.244 (6) & +0.178 {\scriptsize[$-$0.037, +0.392]} \\
       & 10 & 3.051 (6) & 2.915 (6) & \textbf{+0.136 {\scriptsize[+0.060, +0.213]}} \\
      \bottomrule
    \end{tabular}
\end{table}

\section{Baseline reproduction}\label{sec:app-baselines}

Each baseline uses its authors' published settings and is adapted to our settings where necessary. Appendix~\ref{sec:app-hyperparameters} lists every setting, and $^{\dagger}$ marks each setting that differs from the paper's for a reason other than our tournament format. Before a baseline entered our tournaments, we checked our implementation against its own paper by reproducing it with different seeds against a reported result where the paper has one in a comparable setting, and otherwise against the paper's settings. Each check and its pass rule were written down before the checks were run, except the post hoc COALA-PG* diagnostic described below. Tables~\ref{tab:app-fidelity}, \ref{tab:app-fidelity-2} and~\ref{tab:app-fidelity-coin} give every check, its target, our result and the outcome. Below, we report the flagged and excluded baselines.

\paragraph{Flagged baselines.}
SHAPER passes its checks in IMP but falls slightly short in IPD (three-seed mean $2.97$ against a bar of $3.15$, and one of three seeds passed). The paper reports SHAPER only in its own $100\times100$ format, so this check asks whether training in our $50\times20$ format recovers at least half of the reward that a model trained in the paper's format loses in ours: the bar is the midpoint of that model's reward in our format ($2.67$) and in its own ($3.63$). The IMP bars are set in the same way. We enter the model trained with the budget of the authors' configuration files (Table~\ref{tab:hp-shaper-matrix}) and flag its IPD result.
M-FOS passes its Coin Game check on two of three seeds and its returns exceed the paper's, and is flagged as stronger than in its paper. The check used the authors' four-action game. The entrant, trained in the same way in our five-action game and not gated, returns more still.
Reciprocator fails our checks with its paper's formula but passes with its released code, which we enter. Its Coin Game check is uninformative, because in one of the two control runs, two naive learners without a Reciprocator also exceeded the bar. Its shaping also needs more episodes than our 16-episode lifetimes (in the paper the effect appears after about 25 episodes), so it is flagged in Coin Game. We also scored Coin Game without it, and the order of the other four entrants is unchanged (SHAPER $0.231$, M-FOS $0.070$, \method{} $-0.087$, PPO naive learner $-0.472$).

\paragraph{COALA-PG*.}
Our reconstruction of COALA-PG passes the check on the pure-shaping experiment of its paper (Fig.~5A), learning faster than the paper's curve. The entrants use one optimiser step per iteration, which a speed check found to match the paper's pace. It does not reproduce the mixed-pool training (Fig.~5C) either in the check or in the post hoc diagnostic with the paper's optimiser step count. COALA-PG* therefore plays only IPD, trained in the pure-shaping setting. The asterisk marks it as our reconstruction. There is no released code for the method.

\paragraph{Excluded baselines.}
Advantage Alignment has no released code for its IPD experiment: the loss in the authors' released code is written for other environments, and a port of it did not reproduce the paper's IPD result. We rebuilt the method on the code base of LOQA, with the \texttt{loaded-aa} loss and the paper's IPD settings, and tried to reproduce Fig.~1(b) from the paper: on 3 seeds with the IPD payoffs used by LOLA, POLA and LOQA, since the paper does not print its own, and on 3 seeds with every payoff shifted by $+1.5$. None of the seeds passed (0 of 6). LOQA reproduced Fig.~1 on 1 of 3 seeds with the settings in its paper and on 0 of 3 with those in its released code. Both methods are therefore excluded from the tournaments.

\paragraph{Other variants.}
M-FOS-H, our matrix-game version of the history-based M-FOS of the paper's Coin Game experiments, has no paper result to check against. The M-FOS meta-self-play curriculum fails its Coin Game check with the authors' code (1 of 3 seeds). The Coin Game version in the self-play comparison (Table~\ref{tab:selfplay_scores} and Figure~\ref{fig:selfplay_curves}) was trained in the same way on our five-action game and was not gated (Appendix~\ref{sec:app-selfplay}).

\begin{table}[H]
\centering
\small
\setlength{\tabcolsep}{3pt}
\caption{Baseline fidelity in the matrix games (I): reproduction checks against each baseline's paper. Target: the target or bar set before the check, and n/r means no target.}
\label{tab:app-fidelity}
\begin{tabular}{p{0.15\textwidth}p{0.2\textwidth}p{0.18\textwidth}p{0.2\textwidth}p{0.09\textwidth}p{0.1\textwidth}}
\toprule
\raggedright Baseline & \raggedright Check & \raggedright Target & \raggedright Ours & \raggedright Outcome & \raggedright Tournament \tabularnewline
\midrule
\raggedright M-FOS, IPD & \raggedright Table 8 (authors' payoffs), 2 seeds & \raggedright M-FOS $-0.51\pm0.1$, NL $-2.14\pm0.2$ & \raggedright M-FOS $-0.476$, $-0.474$, NL $-2.174$, $-2.176$ & \raggedright pass & \raggedright entrant \tabularnewline
\raggedright M-FOS, IMP & \raggedright Table 5, authors' code (B) vs ours at lr 0.002 (C) & \raggedright 0.20, B $\geq0.18$, $|C-B|\leq0.03$ & \raggedright B 0.194, C 0.194 & \raggedright pass & \raggedright entrant (lr 0.002) \tabularnewline
\raggedright M-FOS, Chicken & \raggedright Table 12 & \raggedright M-FOS above NL & \raggedright M-FOS 3.94, NL 0.99 (5/5 seeds) & \raggedright pass & \raggedright entrant \tabularnewline
\raggedright M-FOS, Stag Hunt & \raggedright no paper result & \raggedright n/r & \raggedright M-FOS 3.93, NL 3.95 & \raggedright -- & \raggedright entrant \tabularnewline
\raggedright M-FOS, meta-self-play & \raggedright self-play, $K=50$ & \raggedright n/r & \raggedright IPD 2.988, IMP 0.012 & \raggedright -- & \raggedright self-play only \tabularnewline
\raggedright SHAPER, IPD & \raggedright recovery, 3 seeds & \raggedright 3.15 & \raggedright 2.967 & \raggedright fail & \raggedright entrant (flagged) \tabularnewline
\raggedright SHAPER, IMP & \raggedright recovery, both seats & \raggedright 0.34, 0.52 & \raggedright 0.719, 0.757 & \raggedright pass & \raggedright entrant \tabularnewline
\raggedright SHAPER, Chicken and Stag Hunt & \raggedright training health & \raggedright n/r & \raggedright 6/6 runs healthy & \raggedright pass & \raggedright entrant \tabularnewline
\bottomrule
\end{tabular}
\end{table}

\begin{table}[H]
\centering
\small
\setlength{\tabcolsep}{3pt}
\caption{Baseline fidelity in the matrix games (II), as in Table~\ref{tab:app-fidelity}. Counts are seeds meeting the condition, and $c$ is the payoff-range factor.}
\label{tab:app-fidelity-2}
\begin{tabular}{p{0.15\textwidth}p{0.2\textwidth}p{0.18\textwidth}p{0.2\textwidth}p{0.09\textwidth}p{0.1\textwidth}}
\toprule
\raggedright Baseline & \raggedright Check & \raggedright Target & \raggedright Ours & \raggedright Outcome & \raggedright Tournament \tabularnewline
\midrule
\raggedright POLA-OM* & \raggedright POLA's steps in our payoff units & \raggedright $5/c$, $0.3/c$, $0.1c$ & \raggedright $c = 5, 2, 4, 4$ & \raggedright -- & \raggedright entrant \tabularnewline \raggedright LOLA-OM & \raggedright LOLA's Table 3 (exact LOLA, authors' payoffs) & \raggedright n/r (paper $-1.06$, $-1.98$) & \raggedright LOLA $-1.04$, NL $-1.98$ & \raggedright match & \raggedright entrant \tabularnewline \raggedright POLA-OM* & \raggedright Outer POLA vs authors' code, Fig.~3 and Table 3 & \raggedright n/r & \raggedright within $10^{-9}$, both reproduced & \raggedright match & \raggedright entrant \tabularnewline
\raggedright Reciprocator, IPD & \raggedright shaping gate, 8 seeds & \raggedright $\geq5/8$ & \raggedright paper formula 0/8, released code 8/8 & \raggedright pass (code) & \raggedright entrant (code) \tabularnewline
\raggedright COALA-PG*, pure shaping & \raggedright Fig.~5A & \raggedright return $\geq0.40$ & \raggedright 1.376, 1.401 & \raggedright pass & \raggedright entrant (IPD) \tabularnewline
\raggedright COALA-PG*, mixed pool & \raggedright Fig.~5C & \raggedright recovery $\geq0.5$ & \raggedright 0.035 & \raggedright fail & \raggedright -- \tabularnewline
\raggedright LOQA & \raggedright Fig.~1 policy, 3 seeds & \raggedright $\geq2/3$ & \raggedright paper 1/3, code 0/3 & \raggedright fail & \raggedright excluded \tabularnewline
\raggedright Advantage Alignment & \raggedright Fig.~1(b) policy, 3 seeds & \raggedright P(C$\,|\,$state) bars & \raggedright LOLA payoffs 0/3, shifted 0/3 & \raggedright fail & \raggedright excluded \tabularnewline
\bottomrule
\end{tabular}
\end{table}

\begin{table}[H]
\centering
\small
\setlength{\tabcolsep}{3pt}
\caption{Baseline fidelity in Coin Game, as in Table~\ref{tab:app-fidelity}. M-FOS returns are in the authors' units, summed over 240 paid steps (M-FOS / NL).}
\label{tab:app-fidelity-coin}
\begin{tabular}{p{0.15\textwidth}p{0.2\textwidth}p{0.18\textwidth}p{0.2\textwidth}p{0.09\textwidth}p{0.1\textwidth}}
\toprule
\raggedright Baseline & \raggedright Check & \raggedright Target & \raggedright Ours & \raggedright Outcome & \raggedright Tournament \tabularnewline
\midrule
\raggedright M-FOS, authors' code & \raggedright updates 951--1000, 3 seeds & \raggedright M-FOS $\geq33.2$, NL $\leq-18.5$ in $\geq2/3$ seeds & \raggedright 70.89 / $-75.00$, 2/3 seeds & \raggedright pass (stronger) & \raggedright -- \tabularnewline
\raggedright M-FOS, pax & \raggedright same, not gated & \raggedright n/r & \raggedright M-FOS 113.55, 105.21, 75.94 & \raggedright -- & \raggedright entrant (flagged) \tabularnewline
\raggedright M-FOS, meta-self-play & \raggedright authors' code, self-play, 3 seeds & \raggedright $\geq15.4$ in $\geq2/3$ seeds & \raggedright 8.74, 14.44, 22.42 & \raggedright fail & \raggedright -- \tabularnewline \raggedright M-FOS, meta-self-play, pax & \raggedright self-play, 3 seeds, not gated & \raggedright n/r & \raggedright 29.95, 25.98, $-3.48$ & \raggedright -- & \raggedright self-play only \tabularnewline
\raggedright SHAPER & \raggedright paper's ES budget & \raggedright n/r & \raggedright 3/3 seeds completed & \raggedright -- & \raggedright entrant \tabularnewline
\raggedright Reciprocator & \raggedright NL own-coin rate, 4 seeds & \raggedright $\geq0.52$ in $\geq3/4$ seeds, mean $\geq0.56$, controls $\leq0.52$ & \raggedright 0.617 (4/4), controls 0.473, 0.531 & \raggedright uninformative & \raggedright entrant (flagged) \tabularnewline
\raggedright PPO NL & \raggedright M-FOS's Table 20 & \raggedright n/r & \raggedright settings match & \raggedright -- & \raggedright entrant \tabularnewline
\bottomrule
\end{tabular}
\end{table}

\section{Implementation details of \method{}}\label{sec:app-implementation}

This section gives the details of \method{} that Section~\ref{sec:method} leaves out. Tables~\ref{tab:hp-duos-matrix} and~\ref{tab:hp-duos-coin} list its settings.

\paragraph{Rule inputs.}
After match $k$, the features $x_{k,s}$ of state $s$ are computed from the agent's own record $\mathcal D^i_k$ of the match. The feature $e_s$ is a one-hot code of $s$. The visit share is $v_s=n_s/T$, where $n_s$ is the number of rounds played in $s$. The opponent's smoothed frequency of action~0 is $\hat q_s=(n^0_s+1)/(n_s+2)$, where $n^0_s$ is the number of those rounds in which the opponent played action~0. The agent's probability of action~0 before the update is $p_s=\pi_{\theta_k}(0\mid s)=\sigma(d_{\theta_k}(s))$. Finally, $\bar r_s$ is the agent's mean reward in $s$ divided by $\kappa$, the largest absolute payoff in the agent's payoff matrix $M^i$ ($\kappa=5$ in IPD, $1$ in IMP and $4$ in Chicken and Stag Hunt). A state that was not visited has $v_s=0$, $\hat q_s=1/2$ and $\bar r_s=0$, and still receives a change.

\paragraph{Meta-network.}
The GRU has 32 units and one memory shared by all states. After each match, it reads the five feature vectors $x_{k,s}$, the match's mean reward divided by $\kappa$ and the agent's payoff matrix $M^i/\kappa$ (50 inputs). For each state, one shared MLP with two hidden layers of 64 units maps $x_{k,s}$ and the GRU's memory to a change in $(-1,1)$. All its units, including the output, are tanh. The meta-network has 14{,}881 parameters. Its MLP weights and biases are drawn from $\mathcal U(\pm1/\sqrt{\text{fan-in}})$ and its GRU weights from $\mathcal U(\pm1/\sqrt{32})$. The GRU biases start at 0.

\paragraph{Action symmetry.} The rule is applied twice to every match: once as observed, and once with the two actions relabelled. Relabelling exchanges the states $(0,0)$ and $(1,1)$ and the states $(0,1)$ and $(1,0)$, swaps the rows and columns of the payoff matrix and negates the log-odds. Each view has its own GRU memory, and the change from the relabelled view is mapped back (exchanging the states and negating) and averaged with the other, which gives $y_{k,s}$. The update is therefore the same whichever action is called~0.

\paragraph{Inner network.}
The inner network has two hidden layers of 32 tanh units and 1{,}314 parameters. Its input codes the state with five indicators (the start state, and one indicator per action for the agent's and the opponent's last actions), and its two outputs $z_0,z_1$ give the log-odds $d_\theta(s)=z_0-z_1$. At the start of each lifetime, the network is fitted to initial log-odds drawn independently from $\mathcal N(0,1)$ for each state.

\paragraph{Fitting the targets.} After match $k$, the targets are $d^{\mathrm{clip}}_s=\mathrm{clip}(d_{\theta_k}(s)+y_{k,s},-12,12)$ (in training, with the perturbed change $\tilde y_{k,s}$ in place of $y_{k,s}$). The network is fitted to them with damped Gauss--Newton steps (Section~\ref{sec:agent_update}). Each step is $\delta=J^\top(JJ^\top+\Lambda)^{-1}u$, where $u_s=d^{\mathrm{clip}}_s-d_\theta(s)$, the row of $J$ for state $s$ is $J_s=\partial d_\theta(s)/\partial\theta$, and $\Lambda=\lambda\,\mathrm{diag}(m)^{-1}$. Equivalently, $\delta$ minimises $\sum_s m_s(J_s\delta-u_s)^2+\lambda\|\delta\|^2$. The weight $m_s$ is the number of rounds played in state $s$, or 1 if $s$ was not visited. The damping $\lambda$ is tiny: $10^{-6}$ times the average of $\|J_s\|^2$ over the five states. So $\delta$ is close to the smallest parameter change that matches the targets to first order. Two more steps recompute $J$, $\lambda$ and $u$ at the new parameters. The fit then keeps as $\theta_{k+1}$ whichever of five candidates has the smallest $\max_s|d_\theta(s)-d^{\mathrm{clip}}_s|$: no change, the parameters after each of the three steps, or the midpoint between $\theta_k$ and the parameters after the third step.

\paragraph{Meta-training.}
Each meta-training step draws 32 pairs of initial policies for every opponent type of the mixture and plays each pair four times with independent match and noise draws, giving 128 lifetimes per type, whose average gradients are weighted by $\rho$. The base rule $\eta^{(0)}$ is trained for 800 meta-training steps, and the gradient's global norm is clipped to 1 before each Adam step (learning rate $10^{-3}$). In IMP, the matching and mismatching seats have separate rules. In IMP self-play, a copy of the seat's rule plays the other seat, and the credited mean value is zero.

\paragraph{Gradient estimate.} The four replays share the initial policies and the opponent (in PSRO, also the drawn naive learner, exploiter or fixed strategy). This makes the leave-one-out baseline $b_k$ a close one. Because their match and noise draws are independent, $b_k$ does not depend on the lifetime's own noise, so the estimator stays unbiased. The noise $\sigma\xi_k$ in $\tilde y_k=y_k+\sigma\xi_k$, with $\xi_k\sim\mathcal N(0,I)$ and $\sigma=0.3$, is added to the changes of all five states before the clip. Hence $\bar\pi_\eta(\cdot\mid\mathcal H_k)=\mathcal N(y_k,\sigma^2I)$, where $\mathcal H_k=(M^i,\pi^i_j,\mathcal D^i_j)$ holds the agent's payoff matrix and its policy and trajectory in each match $j\le k$. The values $V^{\mathrm{cr}}_k$, $k=0,\dots,N$, are computed exactly before each match and after the last update. The gradient passes through both GRU memories over the whole lifetime, but not through the game, the opponent's update or the fit.

\paragraph{Naive learners.}
The naive learners in training take one exact gradient step (Equation~\eqref{eq:nl}) on their value $V$ of Equation~\eqref{eq:exact_value} after each match and read \method{}'s exact table. The step size $\alpha$ is 3 in the base stage and is drawn log-uniformly from $[1,10]$ for each opponent in PSRO. The tournament's naive learner steps on the unnormalised value $V/(1-\gamma)$ with step size 1, which is $\alpha=25$ on $V$, so it learns faster than any naive learner seen in training. AllC and AllD have log-odds $8$ and $-8$ in every state.

\paragraph{Training against exploiters.}
Each PSRO round $\ell$ trains a fresh exploiter $\psi^{(\ell)}$ against the latest rule without noise. The exploiter is an MLP with two hidden layers of 32 tanh units (1{,}605 parameters). Before each match $k$, it reads \method{}'s current table (privileged information), its own previous table and $k/N$, and outputs its own table for the match. Its inputs are centred at $1/2$ and scaled by 4, and its outputs are scaled by 4. Its first previous table is $\sigma(\phi)$ with $\phi\sim\mathcal N(0,I)$. It is trained by antithetic evolution strategies to maximise its own horizon-weighted value $\sum_{k=0}^{N}w_kV^{-i}_k$: 150 iterations of 64 directions, each evaluated on 32 lifetimes with common random numbers, with perturbation scale 0.1 and Adam (learning rate 0.03).

\paragraph{Meta-game and training mixture.} The meta-game matrix $U^{(\ell)}$ holds \method{}'s horizon-weighted value $\sum_{k=0}^{N}w_kV^i_k$ without noise, estimated on 512 common lifetimes, for each earlier rule $\eta^{(0)},\dots,\eta^{(\ell-1)}$ against each exploiter $\psi^{(1)},\dots,\psi^{(\ell)}$. The exploiter mixture $\mu^\star$, which minimises the largest value of any earlier rule against it, is found by linear programming in both IPD and IMP. The rule is then trained for 200 meta-training steps against the mixture $\rho_\ell$ of self-play (0.30), naive learners (0.25), exploiters drawn from $\mu^\star$ (0.30) and fixed strategies (0.15). The fixed strategies are drawn uniformly from tit-for-tat, generous tit-for-tat (forgiving with probability $1/3$), win-stay-lose-shift (staying after a payoff above the mean of the payoff matrix), always action~0, always action~1, and a strategy that plays a random near-deterministic table (log-odds $\pm3$) for 1 to 10 matches and then learns naively. Adam's state is reset at the first round and then carried over. After six rounds, the last rule is deployed. The rule mixture of the meta-game is not used.

\paragraph{Exploitability.}
For Tables~\ref{tab:app-psro} and~\ref{tab:app-transfer}, a fresh exploiter is trained for 300 iterations with its own random seed, shared by the rules being compared, and evaluated on 1{,}024 held-out lifetimes of $N=100$ matches with the horizon weights $w_k$. Exploitability is its value minus the value that the reference strategy (the uniform Nash strategy in IMP, tit-for-tat in the other games) obtains against the same rule.

\paragraph{Coin Game rule.}
A pickup scenario $c$ is the pair of the two players' last pickups in the current episode (none, its own coin or the other player's coin). This gives nine scenarios. The agent tracks its scenario from its own observations, and the scenario resets at every episode. The policy network reads the agent's observation and its current scenario. After each episode, the rule computes for each scenario a code of its identity, the share of the episode's paid steps spent in it, the agent's mean reward divided by 2, four pickup rates per paid step (each player taking its own or the other's coin), the mean normalised advantage (clipped to $[-5,5]$) and the mean probability of the agent's chosen actions. A GRU with 32 units reads these statistics, without the identity codes, for all nine scenarios and the episode's mean reward (73 inputs). An MLP with two hidden layers of 64 tanh units maps each scenario's statistics and the GRU's memory to $\alpha_c=\tanh(\cdot)\in(-1,1)$. In training, noise with $\sigma=0.3$ is added to $\alpha_c$ after the tanh.

\paragraph{Coin Game inner update.} The advantage is $\hat A_t=R_t-f_t^\top\beta$, where $R_t$ is the discounted return to the end of the episode ($\gamma=0.96$). The coefficients $\beta$ are fitted by ridge regression (penalty 1) on the previous episode (on the current one at the first update), with 47 features: the cells of the opponent and both coins, the scenario, the remaining time, its product with the scenario and a constant. The advantages are normalised over the batch. The inner update is one Adam step (learning rate 0.05). Its first moment uses the weighted gradient $\sum_t\alpha_{c_t}\hat A_t\nabla_\theta\log\pi_\theta(a_t\mid o_t,c_t)$, and its second moment uses the same gradient with $\alpha\equiv1$. The step is therefore linear in the multipliers of this and earlier updates. With $\alpha\equiv1$ at every update, it is the plain REINFORCE step. Adam's state persists over the 15 updates of a lifetime, and the policy network is re-initialised for each lifetime.

\paragraph{Coin Game meta-training.} The training opponents are self-play, the same learner with $\alpha\equiv1$ (a REINFORCE learner, not the tournament's PPO NL), a scripted greedy agent that moves to the nearest coin, and a scripted agent that moves only to its own coin. Meta-training differs from the matrix games' in four ways. (i) $V^i_k$ is the mean reward per paid step in episode $k$ over the 512 games (a Monte Carlo estimate, not Equation~\eqref{eq:exact_value}), $V^{\mathrm{cr}}_k$ is credited as in the matrix games, and the 16 episodes are weighted equally ($w_k=1/16$, with $N=15$ updates). (ii) Each meta-training step plays 128 antithetic pairs of lifetimes that share every random draw except the sign of the noise on $\alpha$. Here $b_k$ is the mean of $G_k$ over the other pairs against the same opponent type, and only scenarios visited in the episode enter $\log\bar\pi_\eta$. (iii) The $\alpha\equiv1$ learner's learning rate is multiplied by a factor drawn log-uniformly from $[0.5,2]$ for each lifetime. (iv) After 500 base meta-training steps, three PSRO rounds of 150 meta-training steps follow. Each round trains two fresh exploiters against the latest rule: a GRU policy trained by evolution strategies that also reads \method{}'s action probabilities on 16 fixed probe observations (privileged), and a memoryless policy trained by PPO. The meta-game entry is \method{}'s return minus the exploiter's. Exploiters drawn from its minimax mixture make up 30\% of training, and the other opponents keep their proportions in the remaining 70\%. The last rule is deployed.

\section{Hyperparameters and information access}\label{sec:app-hyperparameters} Tables~\ref{tab:hp-DUOS-matrix} and~\ref{tab:hp-DUOS-coin} give the hyperparameters of \method{}. Those of the baselines are in Tables~\ref{tab:hp-mfos-matrix} and~\ref{tab:hp-mfos-coin} (M-FOS), \ref{tab:hp-shaper-matrix} and~\ref{tab:hp-shaper-coin} (SHAPER), \ref{tab:hp-lola-pola} (LOLA-OM and POLA-OM*), \ref{tab:hp-coala} (COALA-PG*), \ref{tab:hp-reciprocator} (Reciprocator) and~\ref{tab:hp-naive} (naive learners). Table~\ref{tab:hp-protocol} gives the tournament protocol, and Table~\ref{tab:app-info-access} the information each method uses at deployment.

\begin{table}[H]
\centering
\small
\caption{Hyperparameters of \method{} in the matrix games.}
\label{tab:hp-duos-matrix}\label{tab:hp-duos-matrix-2}\label{tab:hp-DUOS-matrix}
\begin{tabular}{lr}
\toprule
Hyperparameter & Value \\
\midrule
Rounds per match $T$ & 20 \\
Matches per training lifetime $N$ & 100 \\
Objective horizons $h$ & 10, 25, 50, 100 \\
Discount $\gamma$ & 0.96 \\
Meta-network GRU units & 32 \\
Meta-network MLP hidden units & 64, 64 \\
Inner network hidden units & 32, 32 \\
Noise $\sigma$ on log-odds changes & 0.3 \\
Lifetimes per opponent type per meta-training step & 128 \\
Opponent mixture $\rho_0$ (Self, NL, AllC, AllD) & 0.4, 0.4, 0.1, 0.1 \\
NL step size on $V$ of Eq.~\eqref{eq:exact_value} (base, PSRO) & 3, log-uniform $[1,10]$ \\
Meta learning rate (Adam) & $10^{-3}$ \\
Base meta-training steps & 800 \\
PSRO rounds $\times$ meta-training steps & $6\times200$ \\
PSRO mixture (Self, NL, exploiters, fixed) & 0.30, 0.25, 0.30, 0.15 \\
Exploiter evolution-strategy (ES) iterations, directions & 150, 64 \\
Training seeds & 5 \\
\bottomrule
\end{tabular}
\end{table}

\begin{table}[H]
\centering
\small
\caption{Hyperparameters of \method{} in Coin Game.}
\label{tab:hp-duos-coin}\label{tab:hp-duos-coin-2}\label{tab:hp-DUOS-coin}
\begin{tabular}{lr}
\toprule
Hyperparameter & Value \\
\midrule
Steps per episode & 16 \\
Episodes per lifetime & 16 \\
Parallel games & 512 \\
Inner network hidden units & 32, 32 \\
Adam second moment & from the $\alpha\equiv1$ gradient \\
Advantage $\hat A_t$ & return minus ridge baseline \\
Inner learning rate (Adam) & 0.05 \\
Return discount $\gamma$ & 0.96 \\
Rule GRU units & 32 \\
Rule MLP hidden units & 64, 64 \\
Pickup scenarios & 9 \\
Noise $\sigma$ on multipliers & 0.3 \\
Lifetimes per meta-training step & 256 \\
Opponent mixture (Self, $\alpha\equiv1$ learner, greedy, own-coin) & 0.4, 0.4, 0.1, 0.1 \\
Meta learning rate (Adam) & $10^{-3}$ \\
Base meta-training steps & 500 \\
PSRO rounds $\times$ meta-training steps & $3\times150$ \\
PSRO mixture (Self, $\alpha\equiv1$, greedy, own-coin, exploiters) & 0.28, 0.28, 0.07, 0.07, 0.30 \\ Exploiter GRU units (ES) & 16 \\ Exploiter MLP hidden units (PPO) & 64, 64 \\ Exploiter iterations (ES, PPO) & 150, 150 \\ $\alpha\equiv1$ learner's learning-rate factor (training) & $[0.5,2]$ \\
Deployed multipliers $\alpha_c$ (measured) & 0.99--1.00 \\
Training seeds & 3 \\
\bottomrule
\end{tabular}
\end{table}

\begin{table}[H]
\centering
\small
\caption{Hyperparameters of M-FOS and M-FOS-H in the matrix games. $^{\dagger}$The paper's Table 14 gives $2\times10^{-4}$, at which our IMP runs reach 0.134 (paper 0.20). The authors' naive-learner script uses $2\times10^{-3}$, which reaches 0.194 (Table~\ref{tab:app-fidelity}). The meta-self-play curriculum keeps $2\times10^{-4}$.}
\label{tab:hp-mfos-matrix}
\begin{tabular}{lr}
\toprule
Hyperparameter & Value \\
\midrule
Actor and critic hidden units & 256 \\
M-FOS-H GRU units & 16 \\
Learning rate & $2\times10^{-4}$ \\
Learning rate, IMP & $2\times10^{-3}$$^{\dagger}$ \\
PPO epochs & 4 \\
PPO clip & 0.2 \\
Entropy coefficient & 0.01 \\
Value-loss coefficient & 0.5 \\
Meta discount & 0.99 \\
Meta-updates & 1{,}024 \\
Lifetimes per update & 4{,}096 \\
Meta-steps per lifetime & 100 \\
Inner discount $\gamma$ & 0.96 \\
NL step size on $V/(1-\gamma)$ (25 on $V$) & 1 \\
NL step size on $V/(1-\gamma)$, IMP & 0.1 \\
Training seeds & 5 \\
\bottomrule
\end{tabular}
\end{table}

\begin{table}[H]
\centering
\small
\caption{Hyperparameters of M-FOS in Coin Game.}
\label{tab:hp-mfos-coin}
\begin{tabular}{lr}
\toprule
Hyperparameter & Value \\
\midrule
Convolution channels & 16, 16 \\
Linear and GRU units & 16, 16 \\
Learning rate & $2\times10^{-4}$ \\
PPO epochs & 16 \\
PPO clip & 0.2 \\
Entropy coefficient & 0.01 \\
Value-loss coefficient & 0.5 \\
Meta discount & 0.99 \\
Meta-updates & 1{,}000 \\
Parallel games & 512 \\
Episodes per lifetime (meta-episode) & 16 \\
Training seeds & 3 \\
\bottomrule
\end{tabular}
\end{table}

\begin{table}[H]
\centering
\small
\caption{Hyperparameters of SHAPER in the matrix games. $^{\dagger}$The paper's Table 11 (IPD) lists a GRU of 25 units, 5{,}000 generations and an ES learning rate of 0.01. We use the authors' configuration files (whose GRU size matches the parameter count in the paper's Table 20), as in Coin Game (Table~\ref{tab:hp-shaper-coin}).}
\label{tab:hp-shaper-matrix}
\begin{tabular}{lr}
\toprule
Hyperparameter & Value \\
\midrule
GRU units & 16$^{\dagger}$ \\
ES population & 1{,}000 \\
Generations & 1{,}000$^{\dagger}$ \\
ES learning rate & 0.1$^{\dagger}$ \\
Perturbation scale & 0.04 \\
Opponents per candidate & 10 \\
Matches per lifetime (trial) & 50 \\
Rounds per match & 20 \\
NL learning rate & 1 \\
NL epochs $\times$ minibatches & $2\times4$ \\
NL discount $\gamma$ & 0.96 \\
Training seeds & 3 \\
\bottomrule
\end{tabular}
\end{table}

\begin{table}[H]
\centering
\small
\caption{Hyperparameters of SHAPER in Coin Game. $^{\dagger}$The paper's table lists 0.01. We use the authors' configuration file.}
\label{tab:hp-shaper-coin}
\begin{tabular}{lr}
\toprule
Hyperparameter & Value \\
\midrule
Linear and GRU units & 16, 16 \\
ES population & 4{,}000 \\
Generations & 3{,}000 \\
ES learning rate & 0.1$^{\dagger}$ \\
Perturbation scale & 0.04 \\
L2 penalty & 0.1 \\
Parallel games & 512 \\
Episodes per lifetime (trial) & 16 \\
NL learning rate & 0.005 \\
NL epochs $\times$ minibatches & $2\times8$ \\
Training seeds & 3 \\
\bottomrule
\end{tabular}
\end{table}

\begin{table}[H]
\centering
\small
\caption{Hyperparameters of LOLA-OM and POLA-OM*. Lists give IPD, IMP, Chicken and Stag Hunt. Here $c$ is the payoff-range factor. Step sizes are on the unnormalised value $V/(1-\gamma)$, as in Table~\ref{tab:hp-naive}.}
\label{tab:hp-lola-pola}
\begin{tabular}{lr}
\toprule
Hyperparameter & Value \\
\midrule
LOLA-OM step size & 1 \\
LOLA-OM look-ahead step & 1 \\
LOLA-OM discount $\gamma$ & 0.96, 0.9, 0.96, 0.96 \\
POLA-OM* inner step ($5/c$) & 1, 2.5, 1.25, 1.25 \\
POLA-OM* outer step ($0.3/c$) & 0.06, 0.15, 0.075, 0.075 \\
POLA-OM* KL weight ($0.1c$) & 0.5, 0.2, 0.4, 0.4 \\
POLA-OM* proximal steps (maximum) & 5{,}000 \\
POLA-OM* discount $\gamma$ & 0.96 \\
Opponent-model step size & 0.05 \\
Opponent-model steps (first match, later) & 50, 1 \\
\bottomrule
\end{tabular}
\end{table}

\begin{table}[H]
\centering
\small
\caption{Hyperparameters of COALA-PG* in IPD. $^{\dagger}$The paper's Table 4 gives 4 epochs $\times$ 2 minibatches. A speed check found one step per iteration to match the pace of its Fig.~5A.}
\label{tab:hp-coala}
\begin{tabular}{lr}
\toprule
Hyperparameter & Value \\
\midrule
Hawk width, heads & 32, 2 \\
PPO iterations & 3{,}000 \\
Meta-batch & 128 \\
Inner batch & 16 \\
Matches $\times$ rounds per lifetime (meta-episode) & $20\times10$ \\
Optimiser steps per iteration & 1$^{\dagger}$ \\ Learning rate & $3\times10^{-4}$ \\
PPO clip & 0.2 \\
Value coefficient & 0.5 \\
NL learning rate & 0.005 \\
NL pool size & 10 \\
Training seeds & 5 \\
\bottomrule
\end{tabular}
\end{table}

\begin{table}[H]
\centering
\small
\caption{Hyperparameters of Reciprocator.}
\label{tab:hp-reciprocator}
\begin{tabular}{lr}
\toprule
Hyperparameter & Value \\
\midrule
\multicolumn{2}{@{}l}{\textit{Matrix games}} \\
Reciprocal reward weight & 5.0 \\
Hidden units & 2 \\
Learning rate & 0.005 \\
Discount $\gamma$ & 0.96 \\
PPO epochs, clip, entropy & 10, 0.1, 0.02 \\
Practice matches per update & 2{,}048 \\
Rounds per practice match & 20 \\
\multicolumn{2}{@{}l}{\textit{Coin Game}} \\
Reciprocal reward weight & 1.0 \\
Convolution channels, GRU units & 4, 16 \\
Learning rate & 0.005 \\
PPO epochs, clip, entropy & 40, 0.15, 0.01 \\
Influence-model learning rate & 0.01 \\
Parallel games & 512 \\
\bottomrule
\end{tabular}
\end{table}

\begin{table}[H]
\centering
\small
\caption{Hyperparameters of the naive learners.}
\label{tab:hp-naive}
\begin{tabular}{lr}
\toprule
Hyperparameter & Value \\
\midrule
\multicolumn{2}{@{}l}{\textit{Matrix games (NL)}} \\
Step size on $V/(1-\gamma)$ (25 on $V$ of Eq.~\eqref{eq:exact_value}) & 1 \\
Discount $\gamma$ & 0.96 \\
\multicolumn{2}{@{}l}{\textit{Coin Game (PPO NL)}} \\
Convolution channels & 16, 16 \\ Linear units & 16 \\
Learning rate & 0.005 \\
PPO epochs, clip & 80, 0.2 \\
Entropy, value-loss coefficients & 0.01, 0.5 \\
Discount $\gamma$ & 0.96 \\
\bottomrule
\end{tabular}
\end{table}

\begin{table}[H]
\centering
\small
\caption{Tournament protocol. Intervals for normalised scores are 95\% percentile bootstrap intervals over 2{,}000 resamples (10{,}000 for Table~\ref{tab:selfplay_scores}) of each learned method's training seeds within each game, recomputing scores, ranks and the min--max normalisation in every resample. Lifetimes are not resampled.}
\label{tab:hp-protocol}
\begin{tabular}{lr}
\toprule
Setting & Value \\
\midrule
\multicolumn{2}{@{}l}{\textit{Matrix games}} \\
Matches per lifetime $K$ & 50 \\
Matches per lifetime $K$, secondary & 10 \\
Rounds per match & 20 \\
Lifetimes per pair of training seeds (IMP: per seat order) & 128 \\
Trained seeds (\method{}, M-FOS, M-FOS-H, COALA-PG*) & 5 \\
Trained seeds (SHAPER) & 3 \\
\multicolumn{2}{@{}l}{\textit{Coin Game}} \\
Steps per episode & 16 \\
Episodes per lifetime & 16 \\
Parallel games & 512 \\
Lifetimes per pair of training seeds and seat & 8 \\
Trained seeds & 3 \\
\midrule
Protocol seed & 23962026 \\
\bottomrule
\end{tabular}
\end{table}

\begin{table}[H]
\centering
\footnotesize
\setlength{\tabcolsep}{3pt}
\caption{Information each method uses at deployment.}
\label{tab:app-info-access}
\begin{tabular}{@{}l|p{0.12\linewidth}p{0.1\linewidth}p{0.1\linewidth}p{0.11\linewidth}p{0.11\linewidth}p{0.13\linewidth}@{}}
Method & Own\newline trajectory & Opponent's\newline actions & Opponent's\newline policy & Opponent's\newline payoffs & Assumed\newline learning rule & Extra \\
\hline
\multicolumn{7}{@{}l}{\textit{Matrix games}} \\
\method{} & last match$^{a}$ & observed & no & no & no & own payoffs, own table \\
M-FOS & no & no & table$^{b}$ & no & no & own table \\
M-FOS-H & last match$^{c}$ & observed & no & no & no & own table \\
SHAPER & every round$^{d}$ & observed & no & no & no & none \\
COALA-PG* & every round$^{d}$ & observed & no & no & no & none \\
LOLA-OM & last match & observed$^{e}$ & fitted$^{e}$ & matrix & NL step$^{f}$ & exact values \\
POLA-OM* & last match & observed$^{e}$ & fitted$^{e}$ & matrix & NL step$^{f}$ & exact values \\
Reciprocator & practice only$^{g}$ & in practice$^{g}$ & table$^{b}$ & matrix & no & practice matches$^{g}$ \\
NL & no & no & table$^{b}$ & no & no & exact values \\
\multicolumn{7}{@{}l}{\textit{Coin Game}} \\
\method{} & last episode & no$^{h}$ & no & no & no & none \\
M-FOS & every step & no & no & rewards$^{i}$ & no & none \\
SHAPER & every step$^{d}$ & no & no & no & no & none \\
Reciprocator & every step & given$^{j}$ & no & rewards$^{j}$ & no & none \\
PPO NL & last episode & no & no & no & no & none \\
\end{tabular}
\par\smallskip
\begin{minipage}{\linewidth}\footnotesize
$^{a}$States, own and opponent actions and own rewards. $^{b}$The opponent's last table, supplied after each match (privileged). $^{c}$The 20 joint actions, through a GRU reset every match. $^{d}$Own observation, kept in a recurrent state over the lifetime. $^{e}$Used only to fit a model of the opponent's table. $^{f}$One naive gradient step. $^{g}$2{,}048 unscored practice matches of 20 rounds per update against the supplied table. $^{h}$Not given explicitly: in Coin Game every method sees both players' positions, and \method{}'s rule uses both players' coin pickups, read from its own observations. $^{i}$Both players' last rewards in its view. $^{j}$Given on each paid step.
\end{minipage}
\end{table}

\section{Additional results}\label{sec:app-results}

\paragraph{Self-play variants.}\label{sec:app-selfplay} \method{} (self-play) in Figure~\ref{fig:selfplay_curves} is exploratory and post hoc. It was designed after Table~\ref{tab:selfplay_scores} had been seen, and it is not a tournament entrant. It uses the entrant's rule form, code and settings, but trains only in self-play from the first meta-training step, with no naive learners, fixed strategies or PSRO. Both seats run the same rule, each with its own inner network, and both copies are credited with their mean value. As the games are symmetric, this gives an unbiased estimate of the gradient, with respect to the shared rule, of one copy's expected return, a common-interest objective.

It is trained separately in IPD, Chicken and Stag Hunt for 2{,}000 meta-training steps of 128 lifetimes of $N=100$ matches, and in Coin Game for 950 meta-training steps of 256 lifetimes of 16 episodes in 512 games (the entrant's totals, $800+6\times200$ and $500+3\times150$), with 5 and 3 seeds. IMP is not trained, since the mean value, and hence the gradient, is zero there.

\paragraph{M-FOS curriculum.} M-FOS (meta-self-play) instead follows its authors' curriculum \citep{lu_model-free_2022}: two separately initialised learners, each trained on its own reward, play each other with probability $1-\lambda$, and otherwise each plays its own batch of naive learners. The paper describes the schedule only qualitatively. In the authors' code, which we follow, $\lambda$ starts at 1 and falls by 0.0015 per update in IPD (1{,}024 updates of 4{,}096 lifetimes of 100 matches) and by 0.005 in Coin Game (1{,}000 updates of one lifetime of 16 episodes in 512 games). It reaches 0 at update 667 and 200 respectively, after which training is pure self-play. In the runs shown, 66--68\% (IPD) and 89--91\% (Coin Game) of updates were self-play, and the last naive-learner update came at update 619--650 and 169--195.

\paragraph{Comparison.} Counted in self-play lifetimes, M-FOS (meta-self-play) trains on about 11 times as many as \method{} (self-play) in IPD (about 2.8 million vs.\ 256{,}000), while in Coin Game \method{} (self-play) trains on about 270 times as many (243{,}200 vs.\ about 900). In IPD, M-FOS observes both players' tables, sets its policy directly and samples its table from its learned Gaussian at every match, as in the authors' code. In both games, its self-play scores pair the two co-trained learners of one run. On the scale of Table~\ref{tab:selfplay_scores}, \method{} (self-play) scores 1.019 in IPD and 1.257 in Coin Game, against 1.028 and 1.248 for M-FOS (meta-self-play). Figure~\ref{fig:selfplay_curves} therefore compares two methods, each trained for self-play with its own recipe and budget, not identical training regimes.

\begin{table}[H]
\centering
\small
\caption{What PSRO adds: the deployed rule against its pre-PSRO base (control) on the same 5 seeds. $d$: after minus before, with a paired 95\% $t$-interval. (a) IMP head-to-head margin after 50 matches. (b) Exploitability: a fresh exploiter's value against the rule minus the value that a reference strategy (IPD: tit-for-tat, IMP: the Nash player) obtains against the rule. Lower is better. IMP has separate rules for the two seats, so it has 10 runs (2 seats $\times$ 5 seeds). Its interval pairs the 5 seeds, each averaged over its two seats.}
\label{tab:app-psro}
\textit{(a) IMP head-to-head margin, $K = 50$}\par\smallskip
\begin{tabular}{lrrlc}
\toprule
Opponent & Control & PSRO & $d$ {\scriptsize[95\% CI]} & Seeds PSRO better \\
\midrule
M-FOS & $-$0.020 & +0.124 & +0.144 {\scriptsize[+0.057, +0.231]} & 5/5 \\
M-FOS-H & $-$0.199 & +0.026 & +0.225 {\scriptsize[+0.145, +0.305]} & 5/5 \\
SHAPER & $-$0.092 & +0.019 & +0.111 {\scriptsize[+0.065, +0.158]} & 5/5 \\
\midrule
Pairings won (all 6 IMP opponents) & 3/6 & 6/6 & & \\
\bottomrule
\end{tabular}
\par\medskip
\textit{(b) Exploitability}\par\smallskip
\begin{tabular}{lrrrlc}
\toprule
Game & Runs & Control & PSRO & $d$ {\scriptsize[95\% CI]} & Runs lower \\
\midrule
IMP & 10 & 0.376 & 0.144 & $-$0.232 {\scriptsize[$-$0.292, $-$0.171]} & 10/10 \\
IPD & 5 & 0.106 & 0.078 & $-$0.029 {\scriptsize[$-$0.114, +0.057]} & 3/5 \\
\bottomrule
\end{tabular}
\end{table}

\begin{table}[H]
\centering
\small
\caption{Zero-shot transfer: the IPD rule played zero-shot compared with rules trained in Chicken and Stag Hunt, each entered in place of it in the same field, after 50 matches (5 seeds, in the field without SHAPER of Table~\ref{tab:shaper-original-vs-extended}, with the trained rules using the IPD recipe, not retuned). $d$: trained minus zero-shot, with a paired 95\% $t$-interval. Seeds: seeds on which the trained rule is better. Self-play return: same-seed pairs from the same tournament runs. Exploitability is as in Table~\ref{tab:app-psro}, with tit-for-tat as the reference. A higher score can come with a lower rank because the other entrants' scores also change: the Chicken-trained rule yields to Straight, which raises M-FOS's score from 1.60 to 2.31 and POLA-OM*'s from 1.60 to 2.20.}
\label{tab:app-transfer}
\begin{tabular}{llrrlc}
\toprule
Game & Measure & Trained & Zero-shot & $d$ {\scriptsize[95\% CI]} & Seeds \\
\midrule
Chicken & Tournament score & 2.105 & 1.810 & +0.295 {\scriptsize[+0.218, +0.371]} & 5/5 \\
 & Rank (of 6) & 3 & 1 & -- & -- \\
 & Own return vs M-FOS & 0.979 & 0.108 & +0.871 {\scriptsize[+0.857, +0.886]} & 5/5 \\
 & Margin vs M-FOS & $-$2.893 & $-$0.195 & $-$2.697 {\scriptsize[$-$2.753, $-$2.642]} & 0/5 \\
 & Self-play return & 2.903 & 2.920 & $-$0.017 {\scriptsize[$-$0.033, $-$0.001]} & 0/5 \\
 & Exploitability & 0.885 & 0.020 & +0.865 {\scriptsize[+0.849, +0.880]} & 0/5 \\
\midrule
Stag Hunt & Tournament score & 3.667 & 3.415 & +0.252 {\scriptsize[+0.190, +0.314]} & 5/5 \\
 & Rank (of 6) & 2 & 4 & -- & -- \\
 & Own return vs M-FOS & 3.930 & 3.706 & +0.224 {\scriptsize[+0.105, +0.343]} & 5/5 \\
 & Margin vs M-FOS & 0.024 & 0.141 & $-$0.117 {\scriptsize[$-$0.158, $-$0.076]} & 0/5 \\
 & Self-play return & 3.905 & 3.803 & +0.103 {\scriptsize[+0.060, +0.146]} & 5/5 \\
 & Exploitability & 0.022 & 0.035 & $-$0.013 {\scriptsize[$-$0.027, +0.001]} & 5/5 \\
\bottomrule
\end{tabular}
\end{table}

\begin{table}[H]
\centering
\small
\caption{Mean head-to-head margin of \method{} (its reward minus the opponent's, per round, or per step in Coin Game) and, in parentheses, the seeds with a positive margin. Matrix games after 50 matches, Coin Game after 16 episodes. --: not in that game.}
\label{tab:app-h2h}
\begin{tabular}{lccccc}
\toprule
Opponent & IPD & IMP & Chicken & Stag Hunt & Coin \\
\midrule
M-FOS & $-$0.268 (0/5) & +0.124 (4/5) & $-$0.195 (0/5) & +0.141 (5/5) & $-$0.216 (0/3) \\
M-FOS-H & $-$0.260 (0/5) & +0.026 (3/5) & -- & -- & -- \\
SHAPER & $-$0.201 (0/5) & +0.019 (4/5) & $-$0.206 (0/5) & +0.120 (5/5) & $-$0.500 (0/3) \\
COALA-PG* & $-$0.017 (2/5) & -- & -- & -- & -- \\
LOLA-OM & $-$0.182 (0/5) & +0.313 (5/5) & $-$0.102 (0/5) & +0.092 (5/5) & -- \\
POLA-OM* & $-$0.197 (0/5) & +0.004 (4/5) & $-$0.219 (0/5) & +0.065 (5/5) & -- \\
NL (Coin: PPO NL) & +3.068 (5/5) & +0.385 (5/5) & +1.769 (5/5) & +0.869 (5/5) & +0.446 (3/3) \\
Reciprocator & +0.710 (5/5) & -- & +0.647 (5/5) & +0.757 (5/5) & +0.935 (3/3) \\
\midrule
Pairings won & 2/8 & 6/6 & 2/6 & 6/6 & 2/4 \\
\bottomrule
\end{tabular}
\end{table}

\begin{table}[H]
\centering
\small
\caption{Frozen-table control in IPD against NL, after 50 matches (5 seeds $\times$ 128 lifetimes). Actual: \method{}'s recorded lifetimes. Frozen: the same NL, from the same first-match policy, against \method{}'s final (match-50) table held fixed from the first match. Exact expectations over 20-round matches. Mean $\pm$ s.e. over seeds. $d$: frozen minus actual, with a paired 95\% $t$-interval (all 5 seeds have the same sign).}
\label{tab:app-frozen}
\begin{tabular}{lccl}
\toprule
Measure & Actual & Frozen & $d$ {\scriptsize[95\% CI]} \\
\midrule
NL cooperation, last match & 0.929 {\scriptsize$\pm$0.010} & 0.141 {\scriptsize$\pm$0.019} & $-$0.788 {\scriptsize[$-$0.823, $-$0.754]} \\
\method{} reward per round, last match & 4.387 {\scriptsize$\pm$0.017} & 1.371 {\scriptsize$\pm$0.051} & $-$3.016 {\scriptsize[$-$3.172, $-$2.861]} \\
\method{} reward per round, lifetime & 3.961 {\scriptsize$\pm$0.024} & 1.421 {\scriptsize$\pm$0.046} & $-$2.540 {\scriptsize[$-$2.721, $-$2.358]} \\
NL reward per round, lifetime & 0.900 {\scriptsize$\pm$0.030} & 1.145 {\scriptsize$\pm$0.023} & +0.245 {\scriptsize[+0.209, +0.281]} \\
\bottomrule
\end{tabular}
\end{table}
\end{document}